\documentclass[lettersize,journal]{IEEEtran}
\usepackage{amsmath,amsfonts}
\usepackage{algorithmic}
\usepackage{algorithm}
\usepackage{array}
\usepackage[caption=false,font=normalsize,labelfont=sf,textfont=sf]{subfig}
\usepackage{textcomp}
\usepackage{stfloats}
\usepackage{url}
\usepackage{verbatim}
\usepackage{graphicx}
\usepackage{cite}
\usepackage{booktabs}
\usepackage{multirow}
\usepackage{float}
\usepackage{multirow}
\usepackage{caption}
\usepackage{xcolor}
\usepackage{colortbl}
\usepackage{hyperref}
\hypersetup{colorlinks=true,linkcolor=blue,urlcolor=blue}

\begin{document}
% 全局设置表格数据
\renewcommand{\arraystretch}{1.2} % 行间距
\setlength{\tabcolsep}{4pt} % 列间距

\newcommand{\question}{%
    \stepcounter{question}%
    \textbf{Q\thequestion:~\ignorespaces}%
}
\newcounter{question}
\setcounter{question}{0}

\definecolor{lightblue}{HTML}{f3f7fc}

\title{PokéVLA: Empowering Pocket-Sized Vision-Language-Action Model with Comprehensive World Knowledge Guidance}

\author{Yupeng Zheng$^{1,2*}$, Xiang Li$^{2,3*}$, Songen Gu$^{4*}$, Yuhang Zheng$^{2,5*}$, Shuai Tian$^{1}$, Weize Li$^{2}$,\\ Linbo Wang$^{1}$, Senyu Fei$^{6}$, Pengfei Li$^{2,3}$, Yinfeng Gao$^{1}$, Zebin Xing$^{1}$, \\ Yilun Chen$^{2}$, Qichao Zhang$^{1}$, Haoran Li$^{1\dagger}$, Wenchao Ding$^{2\dagger}$ \\
\textsuperscript{1}CASIA,
\textsuperscript{2}TARS Robotics,
\textsuperscript{3}Tsinghua University,\\
\textsuperscript{4}Fudan University,
\textsuperscript{5}National University of Singapore,
\textsuperscript{6}Tongji University,\\
        % <-this % stops a space
}



\maketitle

\begin{abstract}
Recent advances in Vision-Language-Action (VLA) models have opened new avenues for robot manipulation, yet existing methods exhibit limited efficiency and a lack of high-level knowledge and spatial awareness. To address these challenges, we propose PokeVLA, a lightweight yet powerful foundation model for embodied manipulation that effectively infuses vision-language understanding into action learning. Our framework introduces a two-stage training paradigm: \textbf{first}, we pre-train a compact vision-language model (PokeVLM) on a curated multimodal dataset of 2.4M samples encompassing spatial grounding, affordance, and embodied reasoning tasks; \textbf{second}, we inject manipulation-relevant representations into the action space through multi-view goal-aware semantics learning, geometry alignment, and a novel action expert. Extensive experiments demonstrate state-of-the-art performance on the LIBERO-Plus benchmark and in real-world deployment, outperforming comparable baselines in success rate and robustness under diverse perturbations. To foster reproducibility and community progress, we will open-source our code, model weights, and the scripts for the curated pre-training dataset. Project page: \url{https://getterupper.github.io/PokeVLA}
\end{abstract}


\section{Introduction}

\begin{figure}[t]
\centering
% \vspace{-1em}
\includegraphics[width=\linewidth]{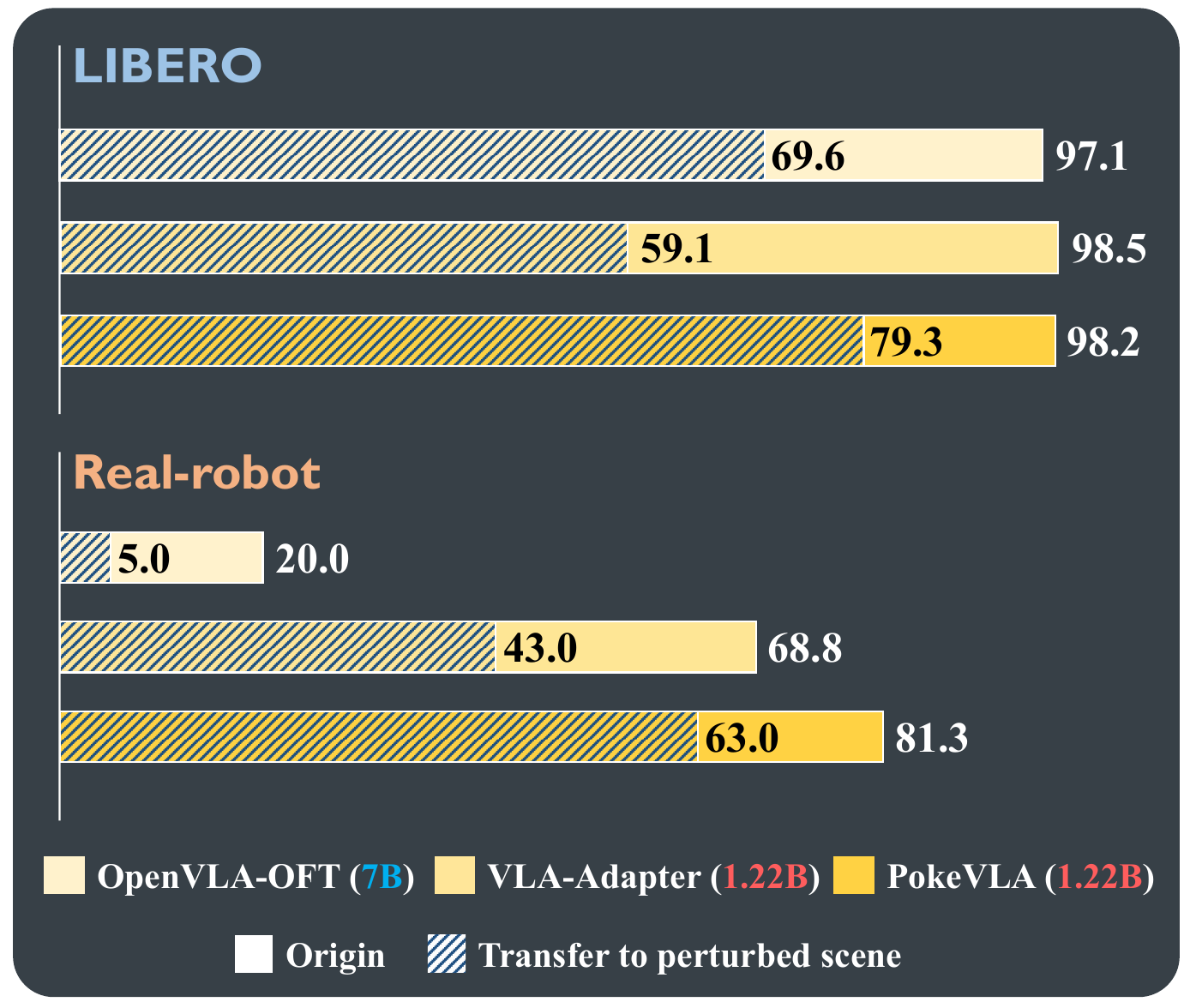}
\caption{\textbf{Success rate in LIBERO benchmark and real-robot tasks.} Our approach effectively leverages pre-trained knowledge to learn representations relevant to the robot manipulation. In both evaluation sets, PokeVLA demonstrated strong performance, maintaining a high success rate even in the presence of significant disturbances in the environment, showcasing its remarkable generalization ability.}
\label{fig:teaser}
\end{figure}

\IEEEPARstart{T}{he} development of foundation models, such as large language models and vision-language models, has opened up a new research paradigm for embodied manipulation. 
How to construct an embodied manipulation foundation model capable of rich perception and precise action planning, such as a Vision-Language-Action (VLA) model, has become a key research direction in the robotics society.

Previous VLA models~\cite{kim2025openvla,kim2025fine} typically rely on pre-training with large-scale robotics manipulation data~\cite{o2024open} to transfer the capabilities of foundation models (e.g., Vision Language Models, VLMs) to an action expert for decoding or generating actions. These methods commonly utilize the hidden-state features from foundation models directly as conditioning inputs for action learning, lacking fine-grained characterization of manipulation-relevant representations. Consequently, these approaches suffer from several bottlenecks, including inefficient action learning and high computational costs.

To address these issues, recent works have explored the critical question of how to effectively infuse the understanding capabilities of VLMs into the action expert to achieve more efficient and accurate action learning. For instance, DreamVLA~\cite{zhang2025dreamvla} establishes a perception-prediction-action loop by forecasting query-based compact world knowledge (dynamic regions, depth, semantics). ReconVLA~\cite{song2026reconvla} guides the action's attention towards perception by reconstructing the target region. VLA-Adapter~\cite{wang2025vlaadapter} employs a bridging attention mechanism to inject the hidden states from each layer of the VLM into the action head. Despite these efforts, existing methods still face three critical challenges:
\textbf{(1) Domain Gap in Pre-trained Knowledge}: A significant domain gap exists when directly applying pre-trained VLMs to embodied manipulation, as their general-purpose knowledge is often misaligned with the specific requirements of robotic tasks.  
\textbf{(2) Lack of Multi-View Spatial Consistency}: The absence of consistent spatial information leads to insufficient generalization to high-level language instructions involving absolute or relative positions.
\textbf{(3) Absence of High-Level Knowledge Prediction}: There is a lack of fine-grained guidance toward the manipulation target, stemming from an inability to predict high-level, task-relevant knowledge.

To overcome these challenges, we introduce PokeVLA, a foundation model for embodied manipulation designed to be (1) lightweight, (2) endowed with rich embodied knowledge, such as target localization, spatial awareness, and affordance prediction, and (3) capable of effectively injecting manipulation-relevant representations into the action learning process. To this end, we devise a two-stage training framework. In the first stage, we pre-train an embodied vision-language foundation model using a large-scale, curated multimodal embodied dataset. In the second stage, we learn manipulation-relevant representations through multi-view semantic learning of the manipulation targets and geometric feature alignment, and then employ action queries to efficiently inject these representations into the action space.

Specifically, our approach entails the following:
 \textbf{(1)} We construct a comprehensive visual-language training dataset of approximately 2.4 million samples, curated from open-source data and simulators, covering four categories: general-purpose visual question answering (VQA), spatial grounding, affordance learning, and embodied reasoning.
 \textbf{(2)} We pre-train a tiny-scale VLM, named PokeVLM, based on the Prismatic-VLM~\cite{karamcheti2024prismatic} framework, which enhances the understanding of embodied scenarios and the spatial reference capabilities while preserving general visual-language capabilities.
 \textbf{(3)} We introduce a learnable special token to learn consistent semantic segmentation of the manipulation targets from both wrist and base camera views, which guides the embodied manipulation.
 \textbf{(4)} Inspired by techniques like Spatial Forcing~\cite{li2025spatial}, we incorporate a feed-forward geometry foundation model to learn multi-view geometry from wrist and base perspectives and distill it into the vision–language model, thereby enhancing its spatial awareness.
 \textbf{(5)} We employ a cross-attention mechanism where action queries aggregate visual features from the VLM's last layer, the learnable special token, and other relevant information. This fused representation is then injected into the action expert.
 
To validate the capability of PokeVLA, we conduct comprehensive evaluations on multiple simulation benchmarks and real-world scenarios. Despite utilizing only a tiny-scale VLM, PokeVLA demonstrates robust performance across various scenarios:
On the large-scale simulation benchmark LIBERO-Plus~\cite{fei2025libero}, PokeVLA achieves state-of-the-art (SOTA) performance. When trained on the LIBERO-Plus dataset, PokeVLA outperforms the baselines OpenVLA-OFT~\cite{kim2025fine} and VLA-Adapter~\cite{wang2025vlaadapter} by 4.0\% and 2.5\% in total success rate, respectively. Furthermore, we conduct generalization experiments on LIBERO-Plus, where the model is trained solely on the original LIBERO~\cite{liu2023libero} dataset and then directly evaluated on diverse environmental variations and perturbations. As shown in Fig.~\ref{fig:teaser}, in this transfer setting, PokeVLA surpasses OpenVLA-OFT and VLA-Adapter by 9.7\% and 20.2\% in average success rate, showcasing its strong generalization capability.
In real-world environments, compared to algorithms of a similar scale, PokeVLA exhibits superior embodied manipulation skills, achieving an 12.5\% improvement in success rate over baseline methods on tasks involving spatial and color referencing. When perturbations are introduced, this performance gap further widens to 20.0\%, highlighting the enhanced robustness of our model.

In summary, our primary contributions are threefold:
\begin{itemize}
    \item  We collect and curate a large-scale embodied multimodal dataset of approximately 2.4 million entries to pre-train a tiny-scale embodied vision-language model. This model acquires rich priors for embodied manipulation while retaining its general visual-language capabilities.
    \item We introduce a novel method for learning manipulation-relevant representations, featuring multi-view consistent learning of the manipulation targets and geometric alignment. This is coupled with a novel action head that efficiently injects these representations into the action learning.
    \item Extensive experiments in both simulation and the real world validate the effectiveness and necessity of our approach in incorporating embodied priors and learning manipulation-relevant representations for robotic manipulation.
\end{itemize}

\section{Related Works}
\subsection{Vision-Language-Action Foundation Models for Robotic Manipulation}

A growing area of interest in robotics is the development of general-purpose models capable of performing a wide range of manipulation tasks across diverse environments. Unlike vision-only controllers~\cite{zhao2023learning, chi2025diffusion, ze2024DP3}, Vision-Language-Action (VLA) models integrate three key modalities within a single architecture: understanding the environment from visual perception, interpreting task descriptions provided in natural language, and generating executable actions based on given inputs. 
Early works such as Octo~\cite{ghosh2024octo} and RT-1~\cite{brohan2023rt} trained transformer-based models from scratch on large-scale robotic demonstration datasets. In recent years, with advancements in Vision-Language Models (VLMs), VLAs have emerged as a significant research direction in robot learning. 
RT-2~\cite{zitkovich2023rt} was among the first to explore the co-fine-tuning of VLMs using web-scale vision-language data alongside robot trajectories, demonstrating strong performance in both accuracy and generalization. 
% To enhance openness and reproducibility, 
OpenVLA~\cite{kim2025openvla} introduced the first open-source VLA model pre-trained on large-scale robot datasets~\cite{o2024open}. 
RoboFlamingo~\cite{li2024vision} demonstrated the feasibility of leveraging a lightly tuned VLM as a perceptual backbone, combined with lightweight action heads, to generate robot actions. 
Building on this foundation, numerous subsequent studies~\cite{black2024pi0, black2025pi05, liu2024towards, fan2025long, bu2025univla} have focused on integrating VLMs with action experts to learn generalizable action and language knowledge. 
For instance, $\pi_0$~\cite{black2024pi0} and $\pi_{0.5}$~\cite{black2025pi05} employ flow-matching decoders to handle complex and high-frequency action generation. 
Inspired by cognitive theories of fast and slow systems, several works have proposed dual-system VLAs, such as GR00T N1~\cite{bjorck2025gr00t} and OpenHelix~\cite{cui2025openhelix}. 
Further efforts have aimed at improving various aspects of model performance. 
OpenVLA-OFT~\cite{kim2025fine} introduced parallel decoding with action chunking, enhancing both inference speed and adaptation efficiency. 
Other studies~\cite{bu2024closed, wen2025diffusionvla, lin2025onetwovla, yang2025instructvla} have focused on strengthening reasoning and planning capabilities, while some~\cite{shukor2025smolvla, wen2025tinyvla, gao2025vla} have explored more efficient architectures. 
By leveraging the robust visual understanding and general reasoning capabilities of pre-trained VLMs, VLAs have achieved impressive results across various manipulation tasks. 
However, recent studies have identified a gap between pre-trained VLMs and embodied environments. 
To better equip VLAs for complex tasks and environments, enhancing their spatial awareness~\cite{zhen20243d, zhang20254d, qu2025spatialvla} and goal-understanding~\cite{zhao2025cot, zhang2025dreamvla, song2026reconvla} capabilities has become an active research focus. 
Building on the OpenVLA-OFT architecture, this work aims to further unleash the intrinsic capabilities of VLMs, enhancing the model’s ability to maintain robust physical-world perception and target-object understanding within a lightweight framework. 
Specifically, we strive to develop a model that is simultaneously: (1) efficient, (2) spatial-aware, and (3) goal-aware.

\begin{figure*}[ht]
\centering
% \vspace{-1em}
\includegraphics[width=\linewidth]{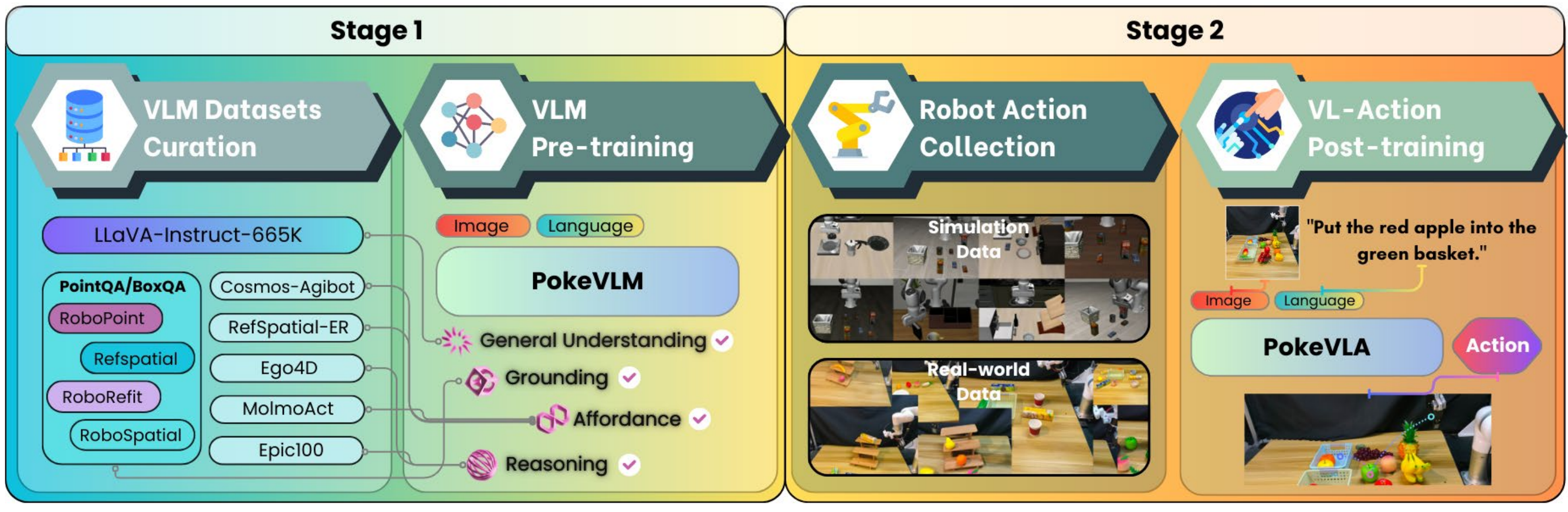}
\caption{\textbf{System overview.} Our system is composed of two stages. The first stage involves pre-training the vision-language model using multimodal embodied data, enhancing its capabilities in understanding and reasoning. In the second stage, we utilize action learning to integrate high-level semantic and spatial information for effective language-instructed manipulation in both simulator and real-world environments.}
\label{fig:system}
\end{figure*}

\textbf{Spatial-Aware VLAs. }
Most existing VLMs are pre-trained solely on 2D image-text data, which inherently limits their accurate understanding of the 3D physical world. To address this gap, a series of studies have sought to empower VLAs with enhanced spatial perception capabilities. 
SpatialVLA~\cite{qu2025spatialvla} introduced 3D positional encodings and adaptive action grids to capture transferable spatial knowledge. 
PointVLA~\cite{li2025pointvla} and GeoVLA~\cite{sun2025geovla} incorporated point embeddings to improve the models' generalizability to unseen tasks. 
BridgeVLA~\cite{li2025bridgevla} takes as input orthographic projection images of 3D point clouds and aligns point clouds across different views by predicting corresponding heatmaps. 
However, these approaches typically require either introducing additional input modalities or relying on external foundation models during inference, leading to increased computational latency. 
Recently, Spatial Forcing~\cite{li2025spatial} leverages foundation models to achieve an implicit spatial alignment of vision embeddings during training, thereby avoiding additional overhead. 
Sharing similar objectives with these efforts, our study aims to enhance spatial comprehension of VLAs in a more systematic manner across three key aspects: (1) we pre-train the VLM backbone on tasks specifically designed for spatial reasoning; (2) during fine-tuning, we learn a multi-view consistent manipulation-relevant representation; and (3) we incorporate an alignment module inspired by \cite{li2025spatial} to further strengthen the model's geometric understanding capabilities.

% \textbf{Efficient VLAs. }

\textbf{Goal-Aware VLAs. }
Several studies have endeavored to enhance VLAs' ability to comprehend target objects and intended actions through auxiliary tasks, which simultaneously improve the interpretability of generated execution. These approaches include generating subgoals or forecasted images~\cite{black2025pi05, zhao2025cot, tian2025predictive}, reconstructing gaze regions of target objects~\cite{song2026reconvla}, or reconstructing dynamic regions~\cite{zhang2025dreamvla}.
Similar to \cite{song2026reconvla}, our work also aims to steer the VLA’s attention toward the target objects. However, instead of generating or reconstructing images, our approach generates semantic segmentation masks of manipulation targets across multiple viewpoints. Inspired by \cite{lai2024lisa}, this strategy not only ensures consistent goal awareness across views but also provides more fine-grained spatial guidance for action generation.

\subsection{Bridging Perception and Action Spaces}
Bridging perception and action spaces constitutes a critical challenge in embodied AI. End-to-end autonomous driving methods~\cite{hu2023planning, jiang2023vad} frequently employ learnable queries as an intermediary bridge between perception and action networks. 
Through cross-attention mechanisms, these queries aggregate crucial information, including panoptic segmentation, online maps, and obstacle attributes, from the perception network, thereby furnishing rich scene representations to facilitate action learning.
However, this problem has received limited investigation in VLA models for robotics manipulation. 
Existing approaches typically extract features from the last layer~\cite{kim2025fine,li2025spatial} of Vision-Language Models (VLMs) to provide perceptual guidance to action experts. 
While recent methods have begun leveraging intermediate-layer~\cite{wang2025vlaadapter} features within VLMs to preserve richer multimodal information, these approaches still primarily transmit raw features, which offer limited refined guidance for action learning.
Drawing inspiration from autonomous driving planning paradigms, we propose to utilize VLMs for learning diverse task-relevant information, such as target regions and spatial awareness. 
By employing learnable queries to aggregate this perception information, closely correlated with embodied tasks, and furnish it to the action policy, we achieve more refined action learning.
\section{System Overview}
As illustrated in Fig.~\ref{fig:system}, our system consists of two stages. 

In the first stage, we collect and integrate multimodal embodied data from open-source datasets for pre-training (Section~\ref{sec: vlm_pretrain_data}). Building upon the prismatic-VLM framework, we construct a vision-language model comprising a Qwen2.5-0.5B language model and DINO-SigLIP dual visual encoders (Section~\ref{sec: vlm_model}). Through pre-training on the integrated embodied data, we obtain a vision-language foundation model endowed with rich knowledge, including the ability of general understanding, spatial grounding, affordance prediction, and embodied reasoning.

\begin{figure*}[t]
\centering
% \vspace{-1em}
\includegraphics[width=0.9\linewidth]{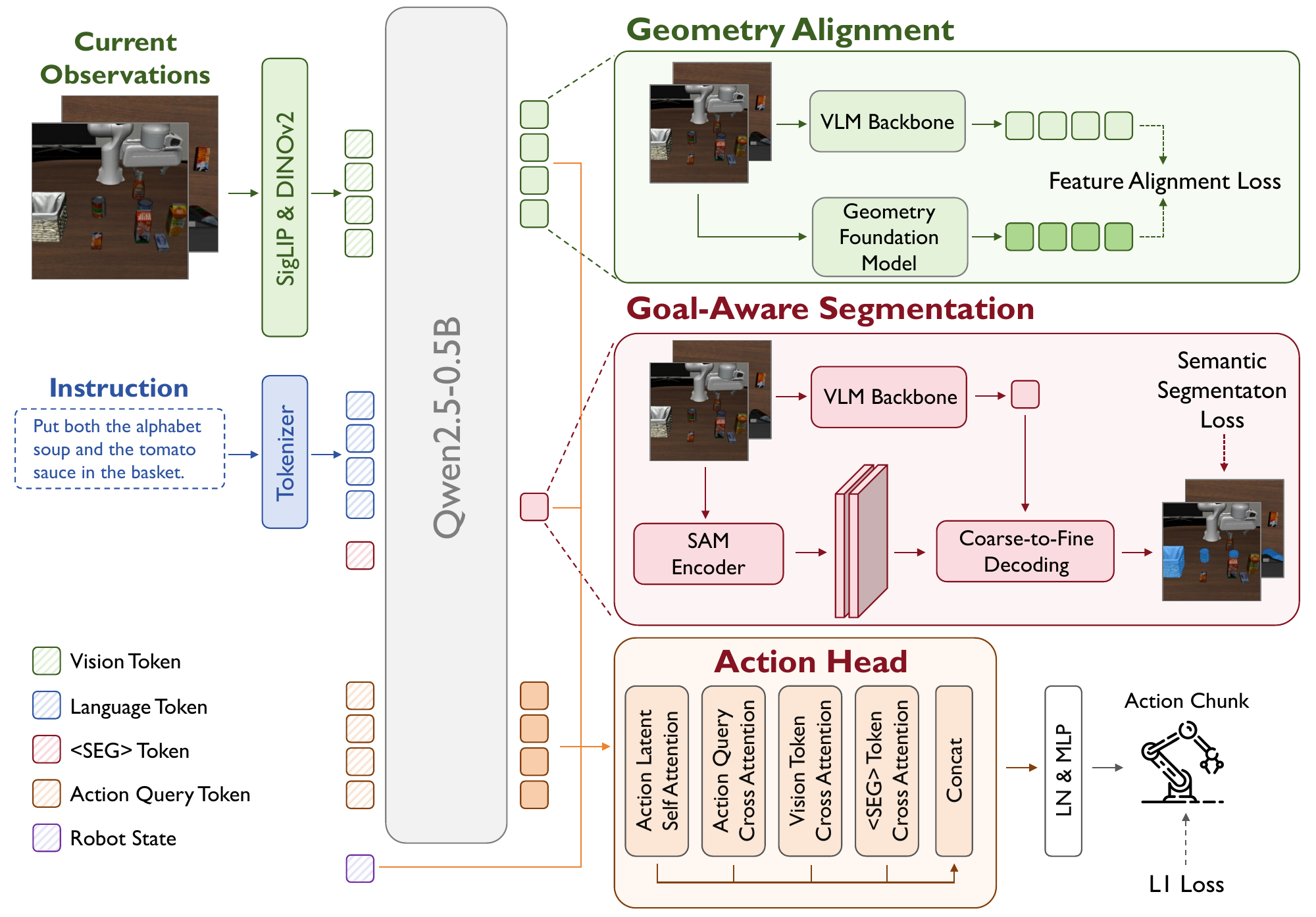}
\caption{\textbf{The architecture of our proposed PokeVLA.} Visual tokens from SigLIP and DINOv2 encoders are fused with language, segmentation, and action query tokens, and then fed into our pre-trained large language model. A geometry alignment module is introduced to align the features of visual tokens with the foundation model, thereby enhancing spatial understanding. Goal-aware segmentation serves as an auxiliary task, training the model to segment target objects across diverse viewpoints to learn a unified semantic scene representation for manipulation. Finally, the action head integrates visual, semantic, language, action and robot state features to predict future action sequences.
}
\label{fig:main}
\end{figure*}

In the second stage, guided by action learning, we acquire high-level semantic information and spatial information relevant to manipulation via learning multi-view manipulation targets (Section~\ref{sec: goal}) and geometry alignment (Section~\ref{sec: geometry}). Finally, we propose a novel action head (Section~\ref{sec: action_head}) that employs action queries to effectively inject these manipulation-relevant representations into the action learning process.
After training with simulation data or data collected from a real-world robot, the model is deployed in simulator and real-world environments to achieve language-instructed embodied manipulation. The architecture of our model is demonstrated in Fig.~\ref{fig:main}.

\begin{table}[t]
\caption{Details of the datasets used in pre-training.}
\label{tab:dataset}
\centering
\renewcommand{\arraystretch}{1.4}
\setlength{\tabcolsep}{10pt}
\begin{tabular}{@{}lcl@{}}
\hline
\textbf{Category} & \textbf{Total} & \textbf{Datasets} \\
\hline
\textbf{General} & 665K & 
\begin{tabular}[t]{@{}l@{}}
llava-1.5-instructions~\cite{liu2024improved}
\end{tabular} \\
\hline
\textbf{Reasoning} & 511K & 
\begin{tabular}[t]{@{}l@{}}
Refspatial (2D Reasoning)~\cite{zhou2025roborefer} \\
Cosmos-Reason1-SFT (Agibot)~\cite{azzolini2025cosmos} \\
Cosmos-Reason1-SFT (Holoassist)~\cite{azzolini2025cosmos}
\end{tabular} \\
\hline
\textbf{Grounding} & 694K & 
\begin{tabular}[t]{@{}l@{}}
Refspatial (Simulator)~\cite{zhou2025roborefer} \\
Robopoint~\cite{yuan2024robopoint} \\
Robospatial (Context)~\cite{song2025robospatial} \\
RoborefIt~\cite{lu2023vl}
\end{tabular} \\
\hline
\textbf{Affordance} & 553K & 
\begin{tabular}[t]{@{}l@{}}
HOVA (Ego4D)~\cite{ma2025glover++} \\
HOVA (Epic100)~\cite{ma2025glover++} \\
MolmoAct~\cite{lee2025molmoact}
\end{tabular} \\
\hline
\end{tabular}
\end{table}

\section{VLM Pre-training}

\subsection{VLM Pre-training Data Collection}
\label{sec: vlm_pretrain_data}

To bridge the gap between general Vision-Language Models (VLMs) and robotics manipulation, we construct a large-scale embodied VLM dataset based on open-source data, aiming to enhance the spatial understanding, object grounding, and manipulation affordance comprehension of VLMs in robotics manipulation tasks. Specifically, our pre-training dataset for VLMs comprises four categories of tasks: General Understanding, Grounding, Affordance, and Reasoning, totaling 2.5 million multimodal samples. The composition and distribution of the dataset are shown in Table~\ref{tab:dataset}.
\begin{itemize}
    \item General Understanding: Following prior works~\cite{liu2023llava,karamcheti2024prismatic}, we adopt LLaVA-Instruct-665K~\cite{liu2024improved} as the dataset for training VLMs on general multimodal comprehension. LLaVA-Instruct-665K is a large-scale multimodal dataset specifically designed for vision-language instruction tuning, covering tasks such as general object recognition, image captioning, and optical character recognition.
    \item Grounding: To enhance the spatial awareness and object localization capabilities of VLMs in tabletop embodied tasks, we integrate data corresponding to grounding tasks from the RoboPoint, RefSpatial, RoboRefit, and robospatial datasets. This results in a combined dataset comprising pointQA and boxQA tasks for localizing target objects and spatial regions.
    \item Affordance: To improve the understanding of action affordances (e.g., graspable points) in embodied tasks, we select the Ego4D~\cite{grauman2022ego4d} and Epic100~\cite{damen2022rescaling} subsets from the HOVA-500K~\cite{ma2025glover++} dataset, which provide operation point information from a human first-person perspective when manipulating diverse objects. We also collect manipulation traces as affordacne from MoloAct~\cite{lee2025molmoact} datasets. 
    \item Reasoning: To strengthen the reasoning capabilities of VLMs in embodied tasks, we collect the embodied reasoning subset from the RefSpatial dataset and the agibot subset from the Cosmos dataset.
\end{itemize}

For all the above datasets, we normalize all point coordinates into relative values within the [0,1] range. Incorrect annotations are removed to retain high-quality samples. For embodied datasets with only word-level annotations (e.g., HOVA), we employ templates and large language models to rewrite and enrich the operational language descriptions. Finally, we perform data sampling for each task to balance the distribution across tasks during VLM pre-training. These integration and quality control measures ensure the reliability of our pre-training dataset, enabling general-purpose VLMs to be adapted into specialized cognitive models for robotic manipulation domains.

\subsection{VLM Model}
\label{sec: vlm_model}
We construct a tiny-scale vision-language model based on Prismatic-VLM~\cite{karamcheti2024prismatic}, which consists of a Qwen2.5-0.5B~\cite{yang2024qwen2} language model backbone, a projector composed of MLPs, and a dual-component visual encoder comprising pretrained SigLIP~\cite{zhai2023sigmoid} and DinoV2~\cite{oquab2023dinov2} models. Here, SigLIP extracts high-level open-vocabulary semantic information, while DinoV2 features are incorporated to enhance spatial perception in robotic manipulation tasks~\cite{karamcheti2024prismatic}. Given an input image, the DINO and SigLIP encoders extract visual features separately, which are then concatenated along the channel dimension to form the combined visual representation. The projector takes this visual representation and maps it into the  space of the language model. During VLM training, we perform end-to-end training via next-token prediction using multimodal data pairs collected from diverse internet sources.

\section{VL-Action Post-Training}

\subsection{Preliminary}
We adopt OpenVLA-OFT as the foundation for building our manipulation foundation model.  OpenVLA-OFT follows the Prismatic-VLMs architecture, which consists of an LLaMA2 7B language model backbone and visual encoders based on DINO and SigLIP. Given multi-view images, robot state, and a language instruction as input, the information from each modality is mapped by respective projection networks into input embeddings, including visual embeddings, language embeddings, and robot state embeddings. These embeddings are then concatenated along the sequence dimension and fed into the language model.

Building upon these inputs, OpenVLA-OFT takes empty action embeddings as input and replaces the causal attention mask with bidirectional attention. For action decoding, OpenVLA-OFT employs parallel decoding, which allows predicting all action steps simultaneously in a single forward pass.
Finally, OpenVLA-OFT replaces the output embedding layer of the decoder with an MLP to learn actions in continuous space. Specifically, the hidden states from the last layer are projected by the MLP into continuous action values. The model is trained by minimizing the L1 loss between predicted and ground-truth actions, enabling efficient inference.

\subsection{Goal-Aware Manipulating}
\label{sec: goal}

\textbf{Manipulation Target Segmentation.} Existing VLAs lack high-level semantic guidance related to manipulation and fail to maintain a consistent understanding of target objects across multiple viewpoints when processing multi-view images. To address this limitation, we introduce manipulation target segmentation as an intermediate reasoning step for action prediction. This directs the model’s attention to regions most relevant to the goal, thereby reducing interference from extraneous visual signals and improving robustness. Compared to existing approaches that reconstruct dynamic or gaze regions, semantic segmentation of individual objects or regions offers more focused and fine-grained guidance for action generation.

Moreover, 2D images lack explicit scene structural information. This requires VLAs to integrate multi-view images for spatial reasoning, which significantly hinders their performance on complex tasks. By training the model to predict pixel-level semantic segmentation masks for multiple manipulation targets across different views, this auxiliary task encourages the learning of a unified, cross-view consistent representation. This, in turn, fosters a more coherent understanding of both the environment and the objects for manipulation.

\textbf{Embedding-as-mask Paradigm.} To endow the VLM backbone with novel region segmentation capabilities without compromising its inherent reasoning ability or the embodied knowledge acquired through large-scale pre-training, we follow the methodology of LISA~\cite{lai2024lisa} and simply expand the original vocabulary of the VLM backbone with a special token \texttt{<SEG>}. By constraining the model output to include the \texttt{<SEG>} token, we can extract its corresponding embedding $h_{seg}\in \mathbb{R}^{1\times D}$ from the last-layer hidden states ($D$ denotes the dimension of backbone features). The \texttt{<SEG>} embedding $h_{seg}$ integrates both contextual scene information and details of the target objects, thereby guiding subsequent mask decoders to accurately localize the relevant regions.

Furthermore, unlike LISA—which is designed for monocular images and performs instance- or part-level segmentation—our approach utilizes only this single token to interact with multi-view image inputs and to generate corresponding multi-view segmentation masks. Consequently, it can implicitly encode knowledge about the 3D structure of the scene. Once the \texttt{<SEG>} embedding is obtained, we simultaneously extract dense visual features from the base camera image $I^b$ and the wrist camera image $I^w$ using the image encoder $\mathcal{E}_{img}$ of the 2D segmentation foundation model SAM~\cite{kirillov2023segment}:
\begin{equation}
f_{sem} = \mathcal{E}_{img}([I^b, I^w])\in \mathbb{R}^{V\times D\times H'\times W'},
\end{equation}
where $V$ denotes the number of viewpoints, and $(H', W')$ represents the downsampled resolution of the images.

\textbf{Coarse-to-Fine Decoding.} Since our \texttt{<SEG>} token aggregates semantic information from multiple viewpoints and involves segmenting multiple objects, it places greater demands on our decoding process. To address this challenge, we adopt a coarse-to-fine decoding paradigm, allowing the model to learn this scene representation progressively. In the coarse decoding stage, the \texttt{<SEG>} embedding $h_{seg}$ is used to guide the model in capturing holistic contextual relationships. Specifically, we feed $h_{seg}$ as a text prompt into SAM's prompt encoder $\mathcal{E}_{prom}$ to obtain corresponding sparse embeddings $ f_c^{sparse} \in \mathbb{R}^{V\times1\times D} $ and dense embeddings $f_c^{dense} \in \mathbb{R}^{V\times D\times H'\times W'}$. These are then fed into the coarse mask decoder $\mathcal{D}_c$ together with $f_{sem}$ to produce semantic logits maps $F$ for each viewpoint:
\begin{equation}
F=\mathcal{D}_{c}(\mathcal{E}_{prom}(h_{seg}), f_{sem}),
\end{equation}
where $F \in \mathbb{R}^{V\times H\times W}$, with $(H, W)$ denoting the resolution of the input images.

In the subsequent fine-grained decoding stage, $F$ serves as a mask prompt, which, along with $h_{seg}$, is fed into the fine-grained mask decoder to guide the generation of more refined prediction maps $\hat{M}\in \mathbb{R}^{V\times H\times W}$:
\begin{equation}
\hat{M}=\mathcal{D}_{f}(\mathcal{E}_{prom}(h_{seg}, F), f_{sem}).
\end{equation}

\textbf{Training Objectives.} We utilize sigmoid focal loss and Kullback-Leibler Divergence loss to compute the mask segmentation loss $L_{mask}$, with corresponding loss weights $\lambda_{focal}$ and $\lambda_{KLD}$. Given ground truth $M$, the total segmentation loss can be formulated as:
\begin{equation}
L_{seg}=\lambda_{focal}\textbf{FOCAL}(\hat{M}, M)+\lambda_{KLD}\textbf{KLD}(\hat{M}, M).
\end{equation}

\subsection{Geometry Alignment}
\label{sec: geometry}

By introducing the auxiliary task of manipulation target segmentation, we endow the model with cross-view consistent goal awareness. However, models that rely solely on 2D images still lack precise geometric perception. Previous approaches have attempted to compensate by incorporating additional modalities or foundation models, but this in turn increases inference latency and hinders real-time performance in real‑world deployment. Similar to~\cite{li2025spatial}, we choose to leverage a powerful 3D geometric foundation model VGGT~\cite{wang2025vggt} only during the training phase. To achieve this, we align the last-layer hidden states corresponding to the vision tokens with features extracted from the foundation model. This alignment enables the intermediate representations of the VLA to learn rich structural information about the scene, while avoiding any additional computational overhead during inference.

Specifically, we first employ VGGT to extract target geometric features $f_{geo}\in\mathbb{R}^{(V\times N)\times D'}$ from the multi-view images, where $N$ denotes the number of vision tokens per image and $D'$ represents the dimension of the VGGT features. Then, we use a lightweight projector $\mathcal{P}$ to ensure that the dimension of the visual hidden states $h_{v}$ matches that of the target features. The geometry alignment loss $L_{geo}$ can be formulated as:
\begin{equation}
L_{geo}=\frac{1}{VN}\sum_{v=1}^{V}\sum_{n=1}^{N}[1-\cos\left( \mathcal{P}(h_v),\; f_{geo} \right)],
\end{equation}
where \(\cos(\cdot,\cdot)\) denotes cosine similarity.

\subsection{Action Head: From VLM to VLA}
\label{sec: action_head}

We integrate the geometrically enhanced hidden states $h_{v}$ (which capture 3D spatial structure) and the semantically refined \texttt{<SEG>} embedding $h_{seg}$ (which captures target-aware semantics), along with the action query embeddings $h_q$ output by the VLM backbone and the current robot proprioceptive state $s_t$, as input to the action head. This conditions the action latents $a_t\in \mathbb{R}^{(T\times D_a)\times D}$ to generate a chunk of $T$ future actions $\hat{A}_t$, where $D_a$ denotes the dimension of each atomic action vector.

Specifically, the action head consists of $L$ layers, each comprising four modules responsible for: self-attention over the action latents $a_t$; cross-attention between $a_t$ and the query embeddings $h_q$ and robot state $s_t$; cross-attention between $a_t$ and the visual hidden states $h_v$; and cross-attention between $a_t$ and the \texttt{<SEG>} embeddings $h_{seg}$. The attention process at the $l$-th layer can be formally expressed as:
\begin{equation}
\texttt{att}_l=[\text{SA}(a_t^l), \text{CA}(a_t^l,  [h_q,\text{MLP}(s_t)]), \text{CA}(a_t^l, h_v), \text{CA}(a_t^l, h_{seg})].
\end{equation}
All attention outputs are concatenated and used to compute the updated action latent $a_t^{l+1}$. Finally, $a_t^{L+1}$ is obtained and passed through layer normalization and an MLP to produce the action chunk $\hat{A}_t$.

Given the ground truth actions $A_t$, we employ an L1 loss between the predicted and ground-truth actions to supervise action generation:
\begin{equation}
L_{action}=\textbf{L1}(\hat{A}_t, A_t).
\end{equation}
Therefore, our overall training objective can be formulated as:
\begin{equation}
L=L_{action}+\lambda_{seg}L_{seg}+\lambda_{geo}L_{geo}.
\end{equation}
For training, we set the hyperparameters as follows: 
$\lambda_{focal} = \lambda_{KLD}=1, \lambda_{seg}=0.2, \lambda_{geo}=0.4$.

% \subsection{Overall Training Objective}
\section{Simulation And VLM Evaluation}

\subsection{Experimental Setup}
\subsubsection{Datasets and Benchmark}

We evaluate our method on two widely adopted simulation benchmarks for VLA models: LIBERO~\cite{liu2023libero} and the more challenging LIBERO‑Plus~\cite{fei2025libero}. Both benchmarks consist of four distinct task suites: Spatial, Object, Goal, and Long. LIBERO‑Plus further introduces seven types of realistic perturbations (i.e., camera viewpoint, robot initialization, language instruction, lighting condition, background texture, sensor noise, and object layout) to test model generalization capabilities and robustness under diverse environmental variations.

For the LIBERO benchmark, each task suite contains approximately 500 training samples (Specifically: Spatial—433, Object—456, Goal—436, and Long—389) and 10 test subtask samples. During evaluation, each test sample is executed 50 times, resulting in a total of 500 test runs per suite.
For the LIBERO-Plus benchmark, there are a total of 15,780 training samples (Spatial—3,970, Object—4,342, Goal—4,034 and Long—3,434). Each suite includes approximately 2,500 test subtask samples, and each test sample is executed once without repetition during evaluation.
To accelerate experimentation, in our ablation studies, we sample 80 test samples from each perturbation type within each suite. Consequently, the total test size for the ablation experiments is 560 runs per suite (4 task types × 7 perturbation types × 80 samples).

\begin{table*}[ht]
\caption{Task description for the Where2Place~\cite{yuan2024robopoint}, Refspatial~\cite{zhou2025roborefer} and CV-Bench~\cite{tong2024cambrian} benchmark.}
\label{tab: vlm_metric}
\centering
\renewcommand{\arraystretch}{1.4}
\setlength{\tabcolsep}{6pt}
\begin{tabular}{@{}ll|l@{}}
\hline
\textbf{Benchmark} &  & \textbf{Description} \\
\hline
Where2Place & Point & Locate vacant regions according to the instructions. \\
            & Bbox & Provide the bounding box coordinates for the specified regions based on the instructions. \\
\hline
            & Location & Point out the objects with specific instructions. \\
Refspatial  & Placement & Point out the free space according to the provided relative spatial description. \\
            & Unseen & Given an unseen, compositionally complex instruction, locate the target object in the image. \\
\hline
\multicolumn{2}{@{}l|}{CV-Bench} & Answer the question according to the image and choices. \\
\hline
\end{tabular}
\label{tab:benchmark_description}
\end{table*}

\subsubsection{Metrics}

For VLM performance, Table~\ref{tab: vlm_metric} presents the evaluation content and metrics for each VLM benchmark. All metrics represent accuracy rates, ranging from 0 to 1, where higher values indicate better performance. For VLA performance, the evaluation metric is the success rate (\%) of the VLA policy. 

\begin{table}[t]
\caption{Quantitative comparison on the Refspatial~\cite{zhou2025roborefer}, Where2Place~\cite{yuan2024robopoint}, and CV-Bench~\cite{tong2024cambrian} benchmark. Higher values indicate better performance.}
\centering
\renewcommand{\arraystretch}{1.3}
\setlength{\tabcolsep}{2pt}
\begin{tabular}{@{}l|cc|ccc|c@{}} 
\hline
\multirow{2}{*}{Model} & 
\multicolumn{2}{c|}{Where2Place} & 
\multicolumn{3}{c|}{Refspatial} & 
\multirow{2}{*}{CV-Bench} \\
% \cline{2-6}
& Point & Bbox & Location & Placement & Unseen & \\
\hline
Prismatic-VLM & 0.075 & 0.095 & 0.033 & 0.012 & 0.015 & 0.455\\
\textbf{PokeVLM (Ours)} & 0.163 & 0.194 & 0.260 & 0.180 & 0.169 & 0.531 \\
\hline
\end{tabular}
\label{tab:vlm_benchmark}
\end{table}

\subsubsection{Implement Details}

\textbf{VLM Pre-training:}
We fine-tune the Prismatic VLM base model for two epochs across 8 GPUs, with the vision projector and language model parameters unfrozen. The effective batch size is 128, achieved through gradient accumulation with a per-GPU batch size of 4 and accumulation steps of 4.
For optimization, we use AdamW~\cite{loshchilov2017decoupled} with a base learning rate of 2e-5. The learning rate is scheduled with a linear warm-up for the first 3\% of training steps, followed by a cosine decay. Gradient clipping with a maximum norm of 1.0 is applied to ensure stable convergence. 

\textbf{Post-training on LIBERO and LIBERO-Plus:}
During the post-training phase, we train PokeVLA across 8 GPUs with a per-GPU batch size of 8, resulting in an effective global batch size of 64.
For optimization, we adopt the AdamW optimizer and LoRA scheme ~\cite{hu2022lora}. For training stability, we configure a learning rate of 1e-4 and utilize a cosine annealing learning rate scheduler with warm-up over the first 10\% of training steps.
For the LIBERO benchmark, we adopt the common practice of training the model separately on each of its task suites for 150K steps. For LIBERO-Plus, we conduct mixed training across all four of its suites uniformly, also for a total of 150K steps.
\subsection{VLM Performance}

\begin{figure}[t]
\centering
% \vspace{-1em}
\includegraphics[width=\linewidth]{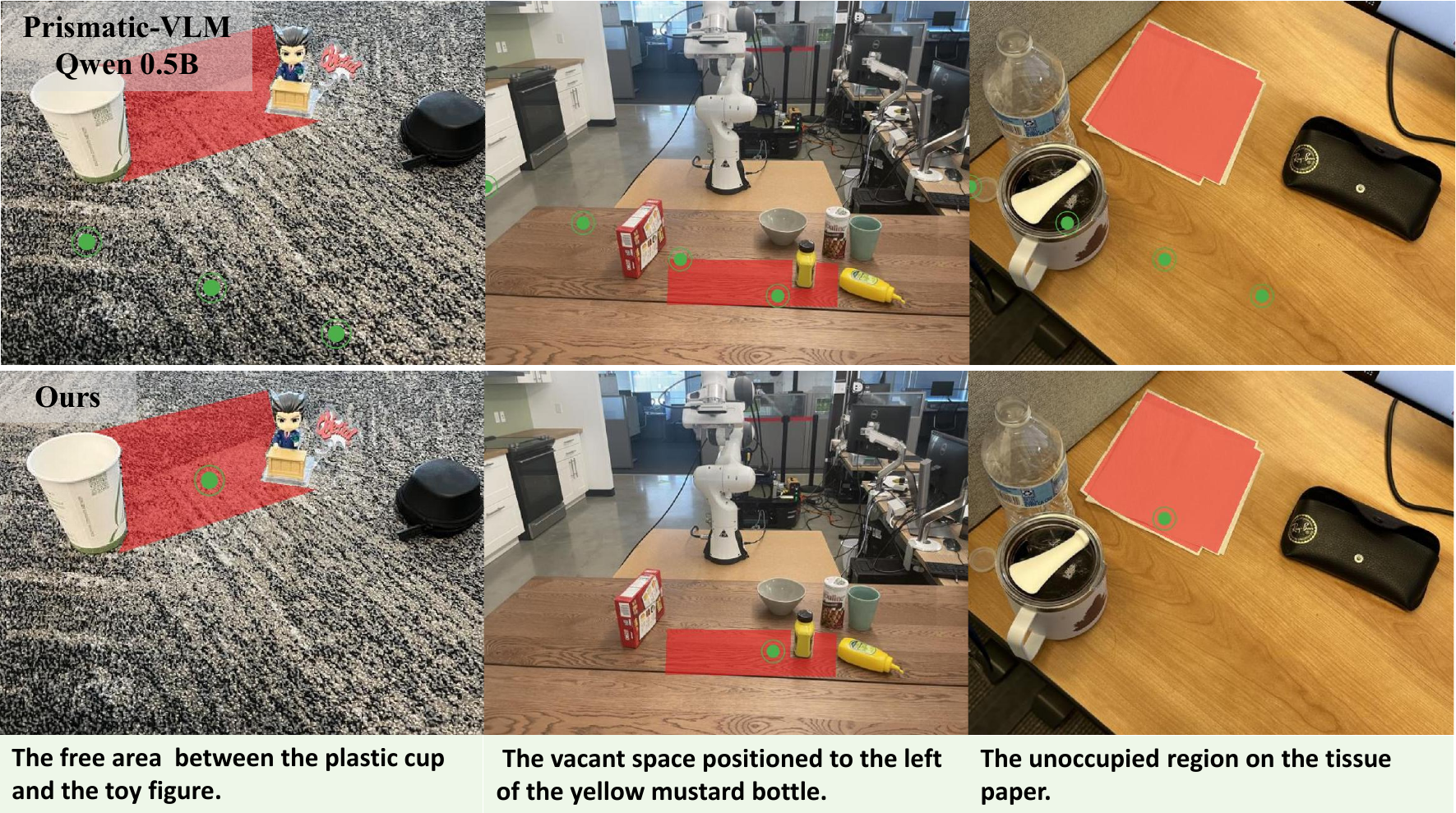}
\caption{\textbf{Visualization of VLM performance on the Where2Place benchmark.} Our model demonstrates superior spatial and semantic comprehension.}
\label{fig:vlm_vis}
\end{figure}

We select three widely used benchmarks, Where2Place~\cite{yuan2024robopoint}, RefSpatial~\cite{zhou2025roborefer}, and CVBench~\cite{tong2024cambrian}, to evaluate PokeVLM on spatial grounding and general understanding tasks. 
The task descriptions of the three benchmarks are shown in Table~\ref{tab:benchmark_description}.
In Table~\ref{tab:vlm_benchmark}, we present the performance of our pre-trained PokeVLM. The experimental results demonstrate that, compared to the original model, PokeVLM achieves a substantial improvement in spatial grounding capability after pre-training, while its general vision-language understanding performance also slightly improves. This indicates that our pre-training not only introduces manipulation-related spatial knowledge but also preserves the inherent vision-language comprehension ability of the VLM.

Fig.~\ref{fig:vlm_vis} further visualizes the results on the Where2Place benchmark, where the red region indicates the spatial area referred to in the language instruction, and the green point represents the model's predicted location. Before pre-training, the model fails to comprehend complex spatial reasoning and repeatedly outputs the same point. After training, the PokeVLM can accurately localize the spatial area described in the instruction. Notably, our training data does not include any samples from Where2Place, demonstrating the generalization ability acquired through our pre-training approach.

\subsection{VLA Simulation Benchmark Performance}

As shown in Table~\ref{tab:libero}, on the standard LIBERO benchmark, PokeVLA achieves a total success rate of 98.2\% with only 1.22 billion parameters, matching the performance of the strongest concurrent methods while substantially outperforming all earlier models of comparable or larger scale. 
Specifically, PokeVLA attains near-saturated performance across all the task suites: Spatial (99.6\%), Object (99.6\%), and Goal (98.4\%). Notably, on the most challenging Long suite, which demand long-horizon planning and goal persistence, it achieves a success rate of 95.2\%, significantly surpassing competing approaches such as CoT-VLA (69.0\%) and WorldVLA (54.1\%), thereby demonstrating its superior capability in complex, long-range manipulation. 

On the more challenging LIBERO-Plus benchmark, we conduct two sets of experiments.

The first set (white rows in Table~\ref{tab:libero_plus}) evaluates models fine-tuned directly on LIBERO-Plus data. Here, PokeVLA exhibits comprehensive and significant performance advantages, achieving a new state-of-the-art of 83.5\% overall success rate. Its robustness is remarkably consistent across diverse perturbation types: it approaches perfection under visual appearance perturbations such as lightning conditions (99.0\%) and background textures (99.3\%); demonstrates strong cross-viewpoint consistency under camera viewpoint shifts (98.2\%), and performs the best under the most difficult robot initialization perturbation, achieving a success rate of 52.9\%. These results collectively underscore the model’s outstanding robustness to real-world distribution shifts.

\begin{table}[t]
\caption{Quantitative comparison on the LIBERO~\cite{liu2023libero} benchmark. Param denotes the scale of the VLM backbone. Concurrent works are highlighted in \textcolor{gray}{gray}.}
\centering
\begin{tabular}{l|c|cccc|c} 
\hline
\multirow{2}{*}{Method} & 
\multirow{2}{*}{Param} & \multicolumn{4}{c|}{Task Suite} & \multirow{2}{*}{Total} \\
 & & Spatial & Object & Goal & Long & \\
\hline
OpenVLA~\cite{kim2025openvla} & 7 & 84.7 & 88.4 & 79.2 & 53.7 & 76.5 \\
OpenVLA-OFT~\cite{kim2025fine} & 7 & 97.6 & 98.4 & 97.9 & 94.5 & 97.1 \\
CoT-VLA~\cite{zhao2025cot} & 7 & 87.5 & 91.6 & 87.6 & 69.0 & 81.1 \\
UniVLA~\cite{bu2025univla} & 7 & 96.5 & 96.8 & 95.6 & 92.0 & 95.2 \\
WorldVLA~\cite{cen2025worldvla} & 7 & 87.6 & 85.2 & 75.1 & 54.1 & 74.8 \\
4D-VLA~\cite{zhang20254d} & 4 & 88.9 & 95.2 & 90.9 & 79.1 & 88.6 \\
SpatialVLA~\cite{qu2025spatialvla} & 4 & 88.2 & 95.2 & 90.9 & 79.1 & 88.6 \\
$\pi_0$~\cite{black2024pi0} & 3 & 96.8 & 98.8 & 95.8 & 85.2 & 94.2 \\
$\pi_0$-FAST~\cite{pertsch2025fast} & 3 & 96.4 & 96.8 & 88.6 & 60.2 & 85.5 \\
SmolVLA~\cite{shukor2025smolvla} & 2.25 & 93.0 & 94.0 & 91.0 & 77.0 & 88.8 \\
GR00t N1~\cite{bjorck2025gr00t} & 2 & 94.4 & 97.6 & 93.0 & 90.6 & 93.9 \\
DreamVLA~\cite{zhang2025dreamvla} & 0.57 & 97.5 & 94.0 & 89.5 & 89.5 & 92.6 \\
\hline
\textcolor{gray}{Spatial Forcing~\cite{li2025spatial}} & \textcolor{gray}{7} & \textcolor{gray}{99.4} & \textcolor{gray}{99.6} & \textcolor{gray}{98.8} & \textcolor{gray}{96.0} & \textcolor{gray}{98.5} \\
\textcolor{gray}{VLA-Adapter~\cite{wang2025vlaadapter}} & \textcolor{gray}{1.22} & \textcolor{gray}{99.6} & \textcolor{gray}{99.6} & \textcolor{gray}{98.2} & \textcolor{gray}{96.4} & \textcolor{gray}{98.5} \\
\hline
\textbf{PokeVLA (Ours)} & 1.22 & 99.6 & 99.6 & 98.4 & 95.2 & 98.2 \\
% \textbf{PokeVLA-InternVL} & 1.1 &  & 98.8 & 96.0 &  &  \\
\hline
\end{tabular}
\label{tab:libero}
\end{table}

The second set (blue rows in Table~\ref{tab:libero_plus}) assesses model transfer performance: all models are fine-tuned exclusively on the standard LIBERO dataset and evaluated without adaptation on LIBERO-Plus. In this demanding direct transfer setting, PokeVLA attains 79.3\% total success rate, substantially outperforming all baselines. It surpasses the strongest competitor, OpenVLA-OFT (69.6\%), by a wide margin. This confirms that the knowledge acquired through our embodied pretraining, goal-aware segmentation, and other tailored training strategies exhibits strong transferability and generalization capability. Notably, PokeVLA’s gains are most pronounced under camera viewpoint shifts (84.7\% over 56.4\%), language instruction variations (84.8\% over 79.5\%), and sensor noise (89.8\% over 75.8\%) scenarios in which many baselines suffer significant degradation. These findings indicate that PokeVLA maintains high robustness against diverse challenging perturbations and generalizes effectively to unseen visual and environmental changes without requiring task-specific fine-tuning.
Moreover, compared to models built upon larger backbones, e.g., OpenVLA-OFT (69.6\%) and $\pi_0$-FAST (61.6\%), our method delivers substantial gains despite its compact parameter budget, highlighting its exceptional parameter efficiency.

% \begin{table*}[ht]
% \caption{Quantitative comparison on the LIBERO benchmark.}
% \centering
% \begin{tabular}{l|c|c|cccc|c} 
% \hline
% \multirow{2}{*}{Method} & \multirow{2}{*}{Venue} & 
% \multirow{2}{*}{Param} & \multicolumn{4}{c|}{Task Suite} & \multirow{2}{*}{Total} \\
%  & & & Spatial & Object & Goal & Long & \\
% \hline
% OpenVLA~\cite{kim2025openvla} & CoRL'24 & 7 & 84.7 & 88.4 & 79.2 & 53.7 & 76.5 \\
% OpenVLA-OFT~\cite{kim2025fine} & RSS'25 & 7 & 97.6 & 98.4 & 97.9 & 94.5 & 97.1 \\
% CoT-VLA~\cite{zhao2025cot} & CVPR'25 & 7 & 87.5 & 91.6 & 87.6 & 69.0 & 81.1 \\
% WorldVLA~\cite{cen2025worldvla} & Arxiv'25 & 7 & 87.6 & 85.2 & 75.1 & 54.1 & 74.8 \\
% 4D-VLA~\cite{zhang20254d} & NeurIPS'25 & 4 & 88.9 & 95.2 & 90.9 & 79.1 & 88.6 \\
% SpatialVLA~\cite{qu2025spatialvla} & RSS'25 & 4 & 88.2 & 95.2 & 90.9 & 79.1 & 88.6 \\
% $\pi_0$~\cite{black2024pi0} & RSS'25 & 3.5 & 96.8 & 98.8 & 95.8 & 85.2 & 94.2 \\
% $\pi_0$-FAST~\cite{pertsch2025fast} & RSS'25 & 3.5 & 96.4 & 96.8 & 88.6 & 60.2 & 85.5 \\
% SmolVLA~\cite{shukor2025smolvla} & Arxiv'25 & 2.25 & 93.0 & 94.0 & 91.0 & 77.0 & 88.8 \\
% Gr00t N1~\cite{bjorck2025gr00t} & Arxiv'25 & 2 & 94.4 & 97.6 & 93.0 & 90.6 & 93.9 \\
% DreamVLA~\cite{zhang2025dreamvla} & NeurIPS'25 & 0.57 & 97.5 & 94.0 & 89.5 & 89.5 & 92.6 \\
% \hline
% \end{tabular}
% \label{tab:libero}
% \end{table*}

\begin{table*}[ht]
\caption{Quantitative comparison on the LIBERO-Plus~\cite{fei2025libero} benchmark. Concurrent works are highlighted in \textcolor{gray}{gray}. \colorbox{lightblue}{Results with a blue background} indicate models trained on unperturbed data and evaluated via direct transfer.}
\centering
\begin{tabular}{l|c|cccc|ccccccc|c} 
\hline
\multirow{2}{*}{Method} & 
\multirow{2}{*}{Param} & \multicolumn{4}{c|}{Task Suite} & \multicolumn{7}{c|}{Perturbation Type} & \multirow{2}{*}{Total} \\
 & & Spatial & Object & Goal & Long & Camera & Robot & Language & Light & Background & Noise & Layout & \\
\hline
\cellcolor[HTML]{f3f7fc}OpenVLA~\cite{kim2025openvla} & \cellcolor[HTML]{f3f7fc}7 & \cellcolor[HTML]{f3f7fc}19.4 & \cellcolor[HTML]{f3f7fc}14.0 & \cellcolor[HTML]{f3f7fc}15.1 & \cellcolor[HTML]{f3f7fc}14.3 & \cellcolor[HTML]{f3f7fc}0.8 & \cellcolor[HTML]{f3f7fc}3.5 & \cellcolor[HTML]{f3f7fc}23.0 & \cellcolor[HTML]{f3f7fc}8.1 & \cellcolor[HTML]{f3f7fc}34.8 & \cellcolor[HTML]{f3f7fc}15.2 & \cellcolor[HTML]{f3f7fc}28.5 & \cellcolor[HTML]{f3f7fc}15.6 \\
\cellcolor[HTML]{f3f7fc}OpenVLA-OFT~\cite{kim2025fine} & \cellcolor[HTML]{f3f7fc}7 & \cellcolor[HTML]{f3f7fc}84.0 & \cellcolor[HTML]{f3f7fc}66.5 & \cellcolor[HTML]{f3f7fc}63.0 & \cellcolor[HTML]{f3f7fc}66.4 & \cellcolor[HTML]{f3f7fc}56.4 & \cellcolor[HTML]{f3f7fc}31.9 & \cellcolor[HTML]{f3f7fc}79.5 & \cellcolor[HTML]{f3f7fc}88.7 & \cellcolor[HTML]{f3f7fc}93.3 & \cellcolor[HTML]{f3f7fc}75.8 & \cellcolor[HTML]{f3f7fc}74.2 & \cellcolor[HTML]{f3f7fc}69.6 \\
\cellcolor[HTML]{f3f7fc}UniVLA~\cite{bu2025univla} & \cellcolor[HTML]{f3f7fc}7 & \cellcolor[HTML]{f3f7fc}55.5 & \cellcolor[HTML]{f3f7fc}36.7 & \cellcolor[HTML]{f3f7fc}40.7 & \cellcolor[HTML]{f3f7fc}39.9 & \cellcolor[HTML]{f3f7fc}1.8 & \cellcolor[HTML]{f3f7fc}46.2 & \cellcolor[HTML]{f3f7fc}69.6 & \cellcolor[HTML]{f3f7fc}69.0 & \cellcolor[HTML]{f3f7fc}81.0 & \cellcolor[HTML]{f3f7fc}21.2 & \cellcolor[HTML]{f3f7fc}31.9 & \cellcolor[HTML]{f3f7fc}42.9 \\
\cellcolor[HTML]{f3f7fc}WorldVLA~\cite{cen2025worldvla} & \cellcolor[HTML]{f3f7fc}7 & \cellcolor[HTML]{f3f7fc}32.5 & \cellcolor[HTML]{f3f7fc}28.6 & \cellcolor[HTML]{f3f7fc}31.8 & \cellcolor[HTML]{f3f7fc}8.2 & \cellcolor[HTML]{f3f7fc}0.1 & \cellcolor[HTML]{f3f7fc}27.9 & \cellcolor[HTML]{f3f7fc}41.6 & \cellcolor[HTML]{f3f7fc}43.7 & \cellcolor[HTML]{f3f7fc}17.1 & \cellcolor[HTML]{f3f7fc}10.9 & \cellcolor[HTML]{f3f7fc}38.0 & \cellcolor[HTML]{f3f7fc}25.0 \\
\cellcolor[HTML]{f3f7fc}$\pi_0$~\cite{black2024pi0} & \cellcolor[HTML]{f3f7fc}3 & \cellcolor[HTML]{f3f7fc}60.7 & \cellcolor[HTML]{f3f7fc}61.4 & \cellcolor[HTML]{f3f7fc}44.9 & \cellcolor[HTML]{f3f7fc}48.4 & \cellcolor[HTML]{f3f7fc}13.8 & \cellcolor[HTML]{f3f7fc}6.0 & \cellcolor[HTML]{f3f7fc}58.8 & \cellcolor[HTML]{f3f7fc}85.0 & \cellcolor[HTML]{f3f7fc}81.4 & \cellcolor[HTML]{f3f7fc}79.0 & \cellcolor[HTML]{f3f7fc}68.8 & \cellcolor[HTML]{f3f7fc}53.6 \\
\cellcolor[HTML]{f3f7fc}$\pi_0$-FAST~\cite{pertsch2025fast} & \cellcolor[HTML]{f3f7fc}3 & \cellcolor[HTML]{f3f7fc}74.4 & \cellcolor[HTML]{f3f7fc}72.7 & \cellcolor[HTML]{f3f7fc}57.6 & \cellcolor[HTML]{f3f7fc}43.4 & \cellcolor[HTML]{f3f7fc}65.1 & \cellcolor[HTML]{f3f7fc}21.6 & \cellcolor[HTML]{f3f7fc}61.0 & \cellcolor[HTML]{f3f7fc}73.2 & \cellcolor[HTML]{f3f7fc}73.2 & \cellcolor[HTML]{f3f7fc}74.4 & \cellcolor[HTML]{f3f7fc}68.8 & \cellcolor[HTML]{f3f7fc}61.6 \\
\hline
\cellcolor[HTML]{f3f7fc}\textcolor{gray}{Spatial Forcing~\cite{li2025spatial}} & \cellcolor[HTML]{f3f7fc}\textcolor{gray}{7} & \cellcolor[HTML]{f3f7fc}\textcolor{gray}{52.9} & \cellcolor[HTML]{f3f7fc}\textcolor{gray}{31.0} & \cellcolor[HTML]{f3f7fc}\textcolor{gray}{28.2} & \cellcolor[HTML]{f3f7fc}\textcolor{gray}{5.4} & \cellcolor[HTML]{f3f7fc}\textcolor{gray}{20.1} & \cellcolor[HTML]{f3f7fc}\textcolor{gray}{13.4} & \cellcolor[HTML]{f3f7fc}\textcolor{gray}{40.9} & \cellcolor[HTML]{f3f7fc}\textcolor{gray}{29.1} & \cellcolor[HTML]{f3f7fc}\textcolor{gray}{33.4} & \cellcolor[HTML]{f3f7fc}\textcolor{gray}{25.7} & \cellcolor[HTML]{f3f7fc}\textcolor{gray}{39.3} & \cellcolor[HTML]{f3f7fc}\textcolor{gray}{29.1} \\
\cellcolor[HTML]{f3f7fc}\textcolor{gray}{VLA-Adapter~\cite{wang2025vlaadapter}} & \cellcolor[HTML]{f3f7fc}\textcolor{gray}{1.22} & \cellcolor[HTML]{f3f7fc}\textcolor{gray}{85.0} & \cellcolor[HTML]{f3f7fc}\textcolor{gray}{46.3} & \cellcolor[HTML]{f3f7fc}\textcolor{gray}{56.0} & \cellcolor[HTML]{f3f7fc}\textcolor{gray}{50.4} & \cellcolor[HTML]{f3f7fc}\textcolor{gray}{36.2} & \cellcolor[HTML]{f3f7fc}\textcolor{gray}{37.9} & \cellcolor[HTML]{f3f7fc}\textcolor{gray}{74.6} & \cellcolor[HTML]{f3f7fc}\textcolor{gray}{70.6} & \cellcolor[HTML]{f3f7fc}\textcolor{gray}{76.1} & \cellcolor[HTML]{f3f7fc}\textcolor{gray}{58.0} & \cellcolor[HTML]{f3f7fc}\textcolor{gray}{69.7} & \cellcolor[HTML]{f3f7fc}\textcolor{gray}{59.1} \\
\hline
\cellcolor[HTML]{f3f7fc}\textbf{PokeVLA (Ours)} & \cellcolor[HTML]{f3f7fc}1.22 & \cellcolor[HTML]{f3f7fc}85.4 & \cellcolor[HTML]{f3f7fc}81.8 & \cellcolor[HTML]{f3f7fc}77.6 & \cellcolor[HTML]{f3f7fc}72.7 & \cellcolor[HTML]{f3f7fc}84.7 & \cellcolor[HTML]{f3f7fc}46.1 & \cellcolor[HTML]{f3f7fc}84.8 & \cellcolor[HTML]{f3f7fc}94.6 & \cellcolor[HTML]{f3f7fc}82.6 & \cellcolor[HTML]{f3f7fc}89.8 & \cellcolor[HTML]{f3f7fc}77.2 & \cellcolor[HTML]{f3f7fc}79.3 \\
% \cellcolor[HTML]{f3f7fc}\textbf{PokeVLA-InternVL} & \cellcolor[HTML]{f3f7fc}1.1 & \cellcolor[HTML]{f3f7fc} & \cellcolor[HTML]{f3f7fc} & \cellcolor[HTML]{f3f7fc} & \cellcolor[HTML]{f3f7fc} & \cellcolor[HTML]{f3f7fc} & \cellcolor[HTML]{f3f7fc} & \cellcolor[HTML]{f3f7fc} & \cellcolor[HTML]{f3f7fc} & \cellcolor[HTML]{f3f7fc} & \cellcolor[HTML]{f3f7fc} & \cellcolor[HTML]{f3f7fc} & \cellcolor[HTML]{f3f7fc} \\
\hline
\hline
OpenVLA-OFT~\cite{kim2025fine} & 7 & 86.1 & 84.5 & 70.7 & 77.7 & 92.8 & 30.3 & 85.8 & 94.9 & 93.9 & 89.3 & 77.6 & 79.5 \\
\hline
\textcolor{gray}{Spatial Forcing~\cite{li2025spatial}} & \textcolor{gray}{7} & \textcolor{gray}{85.9} & \textcolor{gray}{85.5} & \textcolor{gray}{76.5} & \textcolor{gray}{74.9} & \textcolor{gray}{95.2} & \textcolor{gray}{47.9} & \textcolor{gray}{73.5} & \textcolor{gray}{91.2} & \textcolor{gray}{95.6} & \textcolor{gray}{92.2} & \textcolor{gray}{74.8} & \textcolor{gray}{80.5} \\
\textcolor{gray}{VLA-Adapter~\cite{wang2025vlaadapter}} & \textcolor{gray}{1.22} & \textcolor{gray}{83.4} & \textcolor{gray}{82.8} & \textcolor{gray}{77.4} & \textcolor{gray}{80.8} & \textcolor{gray}{96.7} & \textcolor{gray}{42.1} & \textcolor{gray}{71.1} & \textcolor{gray}{96.5} & \textcolor{gray}{97.5} & \textcolor{gray}{96.9} & \textcolor{gray}{74.4} & \textcolor{gray}{81.0} \\
\hline
\textbf{PokeVLA (Ours)} & 1.22 & 87.3 & 86.6 & 80.0 & 80.6 & 98.2 & 52.9 & 71.4 & 99.0 & 99.3 & 94.0 & 77.9 & 83.5 \\
% \textbf{PokeVLA-InternVL} & 1.1 & 88.1 & 81.8 & 72.7 & 75.8 & 92.4 & 44.3 & 74.4 & 98.2 & 95.7 & 94.8 & 65.0 & 79.4 \\
\hline
\end{tabular}
\label{tab:libero_plus}
\end{table*}

\begin{table*}[ht]
\caption{Ablation on each component.}
\centering
\begin{tabular}{ccc|cccc|ccccccc|c} 
\hline
\multirow{2}{*}{Pre-training} & \multirow{2}{*}{Geometry} & \multirow{2}{*}{Goal} & \multicolumn{4}{c|}{Task Suite} & \multicolumn{7}{c|}{Perturbation Type} & \multirow{2}{*}{Total} \\
 & & & Spatial & Object & Goal & Long & Camera & Robot & Language & Light & Background & Noise & Layout & \\
\hline
 & & & 83.2 & 82.9 & 71.2 & 75.5 & 94.7 & 36.3 & 67.8 & 90.0 & 93.8 & 93.1 & 71.9 & 78.2 \\
\checkmark & & & 87.1 & 85.2 & 80.5 & 78.6 & 94.7 & 46.6 & 73.1 & 98.1 & 95.9 & 94.7 & 76.9 & 82.9 \\
 & \checkmark & & 86.4 & 86.2 & 70.0 & 81.4 & 97.5 & 34.7 & 70.0 & 97.8 & 96.9 & 93.1 & 77.2 & 81.0 \\
 & & \checkmark & 83.8 & 85.2 & 77.9 & 82.9 & 98.1 & 39.1 & 76.9 & 99.7 & 98.4 & 96.3 & 68.4 & 82.5  \\
% \checkmark & \checkmark & & 88.6 & 85.4 & 78.9 & 83.0 & 97.8 & 43.8 & 71.9 & 98.4 & 97.8 & 96.9 & 81.3 & 84.0 \\
\checkmark & \checkmark & \checkmark & 88.8 & 88.6 & 81.1 & 83.0 & 97.5 & 55.3 & 72.5 & 99.1 & 99.1 & 96.6 & 77.5 & 85.3 \\
\hline
\end{tabular}
\label{tab:ablation_full}
\end{table*}

\begin{table*}[ht]
\caption{Ablation on pre-training data.}
\centering
\begin{tabular}{l|cccc|ccccccc|c} 
\hline
\multirow{2}{*}{Method} & \multicolumn{4}{c|}{Task Suite} & \multicolumn{7}{c|}{Perturbation Type} & \multirow{2}{*}{Total} \\
 & Spatial & Object & Goal & Long & Camera & Robot & Language & Light & Background & Noise & Layout & \\
\hline
Vanilla & 83.2 & 82.9 & 71.2 & 75.5 & 94.7 & 36.3 & 67.8 & 90.0 & 93.8 & 93.1 & 71.9 & 78.2 \\
\hline
+ Pre-training & 87.1 & 85.2 & 80.5 & 78.6 & 94.7 & 46.6 & 73.1 & 98.1 & 95.9 & 94.7 & 76.9 & 82.9 \\
\ \ \ \textit{w/o} Grounding \ \  & 87.7 & 85.5 & 78.0 & 76.4 & 97.2 & 42.5 & 71.9 & 97.2 & 97.8 & 94.1 & 72.8 & 81.9 \\
\ \ \ \textit{w/o} Affordance \ \  & 86.8 & 85.7 & 76.1 & 77.7 & 96.3 & 39.7 & 77.5 & 93.1 & 96.6 & 91.3 & 76.6 & 81.6 \\
\ \ \ \textit{w/o} Reasoning \ \  & 85.4 & 85.7 & 78.8 & 79.5 & 97.5 & 50.3 & 66.9 & 96.6 & 95.3 & 95.6 & 74.1 & 82.4 \\
\hline
\end{tabular}
\label{tab:ablation_pretraining}
\end{table*}

\subsection{Ablation Study and In-depth Analysis}

In this section, we design a series of experiments to investigate the following questions.

\question \textbf{What is the contribution of each component?}
We performed ablation studies to evaluate the effectiveness of various components within our method, as detailed in Table~\ref{tab:ablation_full}. It can be observed that the individual inclusion of each component yields performance gains over the baseline.

Pre-training results in consistent and significant improvements across all task suites and perturbation types, boosting the total success rate from 78.2\% to 82.9\%. This module is particularly effective in enhancing performance on the Goal suite (increasing from 71.2\% to 80.5\%) and under language instruction perturbations (from 67.8\% to 73.1\%). These results suggest that pre-training on large-scale embodied manipulation data effectively enhances the VLM backbone's semantic understanding of tasks, providing the model with robust general-purpose scene knowledge and a solid foundation for instruction comprehension. Notably, the model's performance under robot initialization perturbations improved markedly (from 36.3\% to 46.6\%), indicating that pre-training significantly mitigates failures caused by variations in the executor's initial pose, thereby enhancing robustness.

Geometry alignment demonstrates a clear improvement in robustness for Long suite (from 75.5\% to 81.4\%) and against various visual perturbations, such as lighting conditions and object layout. This validates that feature alignment with a strong geometric foundation model strengthens the model's understanding of the scene's 3D structure and spatial relationships between objects, leading to more stable performance in complex tasks requiring long-range planning.

When goal-aware segmentation is introduced as an auxiliary task, the success rates under lighting, background, and noise perturbations reached 99.7\%, 98.4\%, and 96.3\%, respectively. These figures are close to or at the state-of-the-art level and represent a significant improvement over the baseline. We attribute this to the module's ability to actively segment the regions most relevant to the manipulation targets, thereby guiding the model's attention to focus on the target objects and automatically filtering out substantial irrelevant visual variations. Crucially, this module provides the most prominent improvement for long-horizon tasks (from 75.5\% to 82.9\%) and camera viewpoint perturbations (from 94.7\% to 98.1\%). This indicates that by steering the model to focus on goal-relevant regions, the module enhances cross-view consistency, ensuring the model maintains a unified target representation during action generation, which is vital for stability in long-horizon tasks.

Together, the full model outperforms any single module configuration, with an increase from 78.2\% to 85.3\% in overall success rate. Its performance under robot initialization perturbations (55.3\%) is considerably higher than using only pre-training or goal-aware segmentation, highlighting the synergistic effect among modules. Pre-training provides rich semantic priors, geometry alignment enhances the accuracy of spatial reasoning, and goal-aware segmentation ensures the cross-view alignment of visual attention with task goals. The combination of these three components enables the model to simultaneously address semantic, geometric, and task-focus challenges, achieving comprehensive and robust performance across diverse task types and severe perturbation conditions.

\begin{figure*}[htbp]
\centering
\includegraphics[width=\textwidth]{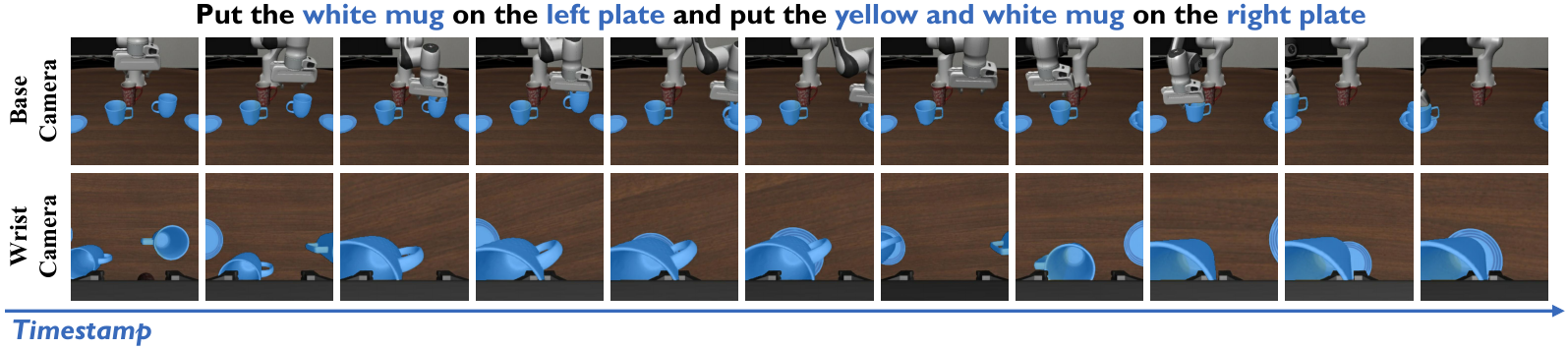}
% \vspace{-2em}
\caption{\textbf{The consistency of goal-aware segmentation results.} Our method maintains consistently high-quality manipulation target segmentation results across long-horizon tasks, demonstrating strong cross-view coherence and temporal consistency.}
\label{fig:consistency}
\end{figure*}

\begin{figure}[t]
\centering
% \vspace{-1em}
\includegraphics[width=\linewidth]{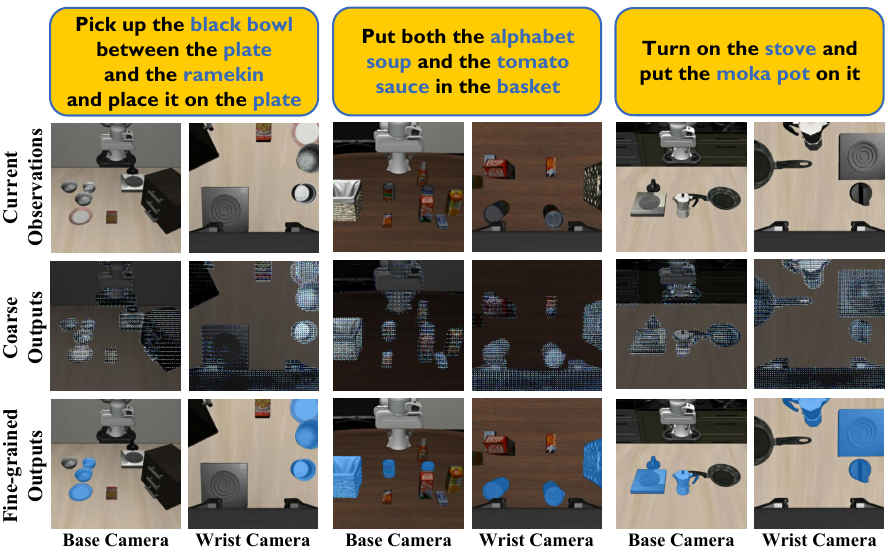}
\caption{\textbf{Visualization of coarse-to-fine decoding paradigm.} It can be observed that our coarse mask decoder captures the spatial relationships among foreground objects in the scene, while the fine-grained mask decoder builds upon this foundation to generate precise semantic segmentation of the manipulation targets for manipulation. Best viewed zoom in.}
\label{fig:coarse-to-fine}
\end{figure}

\question \textbf{What is the contribution of different categories of pre-training data?}
As shown in Table~\ref{tab:ablation_pretraining}, we observe that different categories of pre-training data contribute distinctively yet complementarily to the VLA model's performance. While grounding and affordance data appear more critical to overall performance than reasoning data, the removal of any single component introduces specific capability deficiencies.

The absence of visual grounding data leads to significant performance drop in goal and long-horizon tasks. This indicates that the contextual localization capability—precisely associating language instructions with scene entities—is essential for understanding complex task goals and maintaining consistency during long-horizon execution.

The lack of affordance data severely weakens the model's robustness against robot initialization perturbations (dropping from 46.6\% to 39.7\%). This confirms that explicit learning of object interaction properties directly enhances the adaptability and stability of action generation under variations in the robotic arm's initial pose.

The omission of reasoning data causes a substantial decline in performance under language perturbations (from 73.1\% to 66.9\%). This suggests that data containing causal reasoning and Chain-of-Thought (CoT) traces significantly strengthens the model's ability to parse and generalize diverse, non-standard instructions.

In summary, we argue that an effective embodied pre-training data system must be multi-dimensional: grounding data improves language-scene matching generalization, affordance data enhances action execution robustness, and reasoning data deepens task understanding. Only through the synergy of all three can a VLA model be constructed that is robust across semantic, geometric, and logical dimensions.

\begin{figure}[t]
\centering
% \vspace{-1em}
\includegraphics[width=\linewidth]{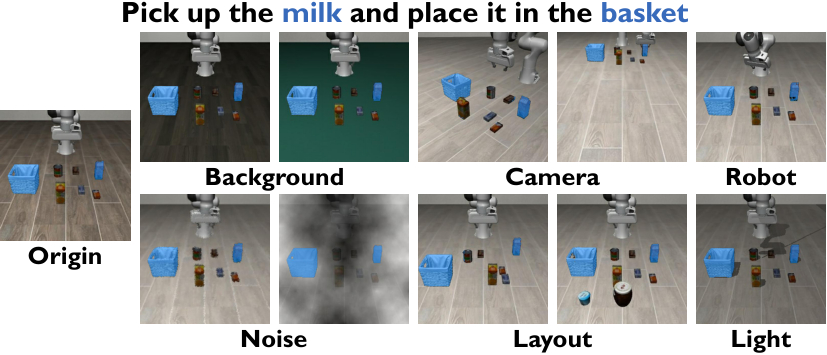}
\caption{\textbf{The robustness of goal-aware segmentation results.} Our method achieves accurate segmentation under a wide range of perturbations.}
\label{fig:robustness}
\end{figure}

\begin{figure*}[ht]
\centering
\includegraphics[width=0.9\textwidth]{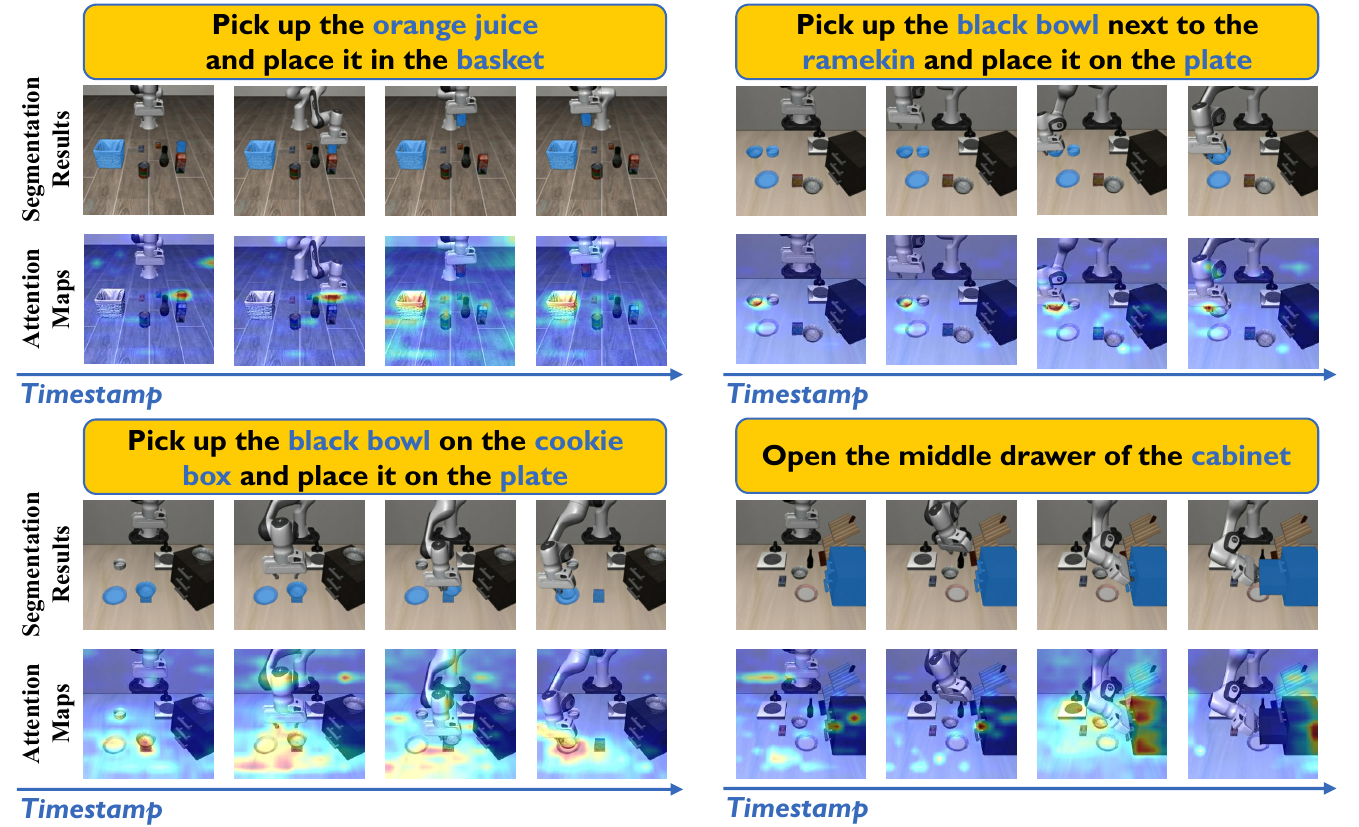}
% \vspace{-2em}
\caption{\textbf{Visualization of attention maps.} Supervised by the goal-aware segmentation as an auxiliary task, our model can precisely guide its attention to the manipulation targets, thereby achieving superior performance across diverse tasks.}
\label{fig:attention}
\end{figure*}

% \question \textbf{What types of knowledge are learned by the different decoders in the goal-aware segmentation task?}
\question \textbf{Within the coarse-to-fine paradigm for goal-aware segmentation, what is the respective role and learned knowledge of each decoding stage?}
In Fig.~\ref{fig:coarse-to-fine}, we visualize the outputs of both stages. It can be seen that while the semantic logit maps generated during the coarse decoding stage exhibit some noise, they successfully segment all foreground objects in the scene. This indicates that the model at this stage has already acquired a preliminary understanding of the spatial relationships and semantic functions of the objects. Furthermore, it is notable that the model exhibits higher attentional response values for manipulation targets, as evidenced by the denser distribution of high-value logits on these objects. This is particularly evident in the column of wrist camera, suggesting that the model has already established a degree of focus on the targets during the first stage.

In the subsequent fine-grained decoding stage, the model aggregates the contextual and structural knowledge learned from the previous stage and the \texttt{<SEG>} token. This allows it to concentrate its attention more intensively on the target objects, yielding high-quality semantic segmentation results.

Moreover, by comparing the base and wrist views, we observe that the model outputs maintain strong cross-view consistency in both the coarse and fine-grained stages. This demonstrates that the model has learned a unified manipulation-relevant representation across both stages.

\question \textbf{How consistent and robust are the representations learned through goal-aware segmentation?}
We visualize the segmentation results on long-horizon tasks in Fig.~\ref{fig:consistency}. Even when individual viewpoints are constrained by a limited field of view (FOV), the model maintains segmentation accuracy over long temporal sequences. This demonstrates that the target representations learned by the model are not only cross-view consistent (as observed in Q3) but also possess high temporal stability. We believe this unified and continuous representation facilitates the model's goal-following capability during the multi-step reasoning process of complex tasks, thereby improving the success rate of task execution.

In Fig.~\ref{fig:robustness}, we present the segmentation results under various perturbation types. It is evident that our method effectively rejects interference from background textures, sensor noise, and lighting conditions. Even under drastic changes in camera viewpoint or the appearance of unseen confounding objects in the scene, the model maintains segmentation accuracy. This reflects the robustness of the learned representations. As discussed in Q1, we posit that this representation guides the model to remain focused on task-relevant objects even in extreme environments, thereby enhancing robustness in confronted settings.

\question \textbf{In what ways do the learned representations guide the process of action generation?}
In Table~\ref{tab:ablation_full}, we quantitatively demonstrated the performance gains attributed to the goal-aware segmentation task. In this section, we further provide a qualitative demonstration through the visualization of attention maps. As can be seen in Fig.~\ref{fig:attention}, this auxiliary task facilitates a tight alignment between the model's attention and the manipulation targets. The attention is even focused on specific affordance regions, such as the rim of the black bowl in the top-right corner and the handle of the middle cabinet drawer in the bottom-right corner, showcasing strong semantic understanding and target localization capabilities. This demonstrates that the manipulation-relevant representations learned through goal-aware segmentation provide highly precise object grounding information, thereby steering the executor to generate high-quality actions.

\begin{figure}[htbp]
\centering
% \vspace{-1em}
\includegraphics[width=\linewidth]{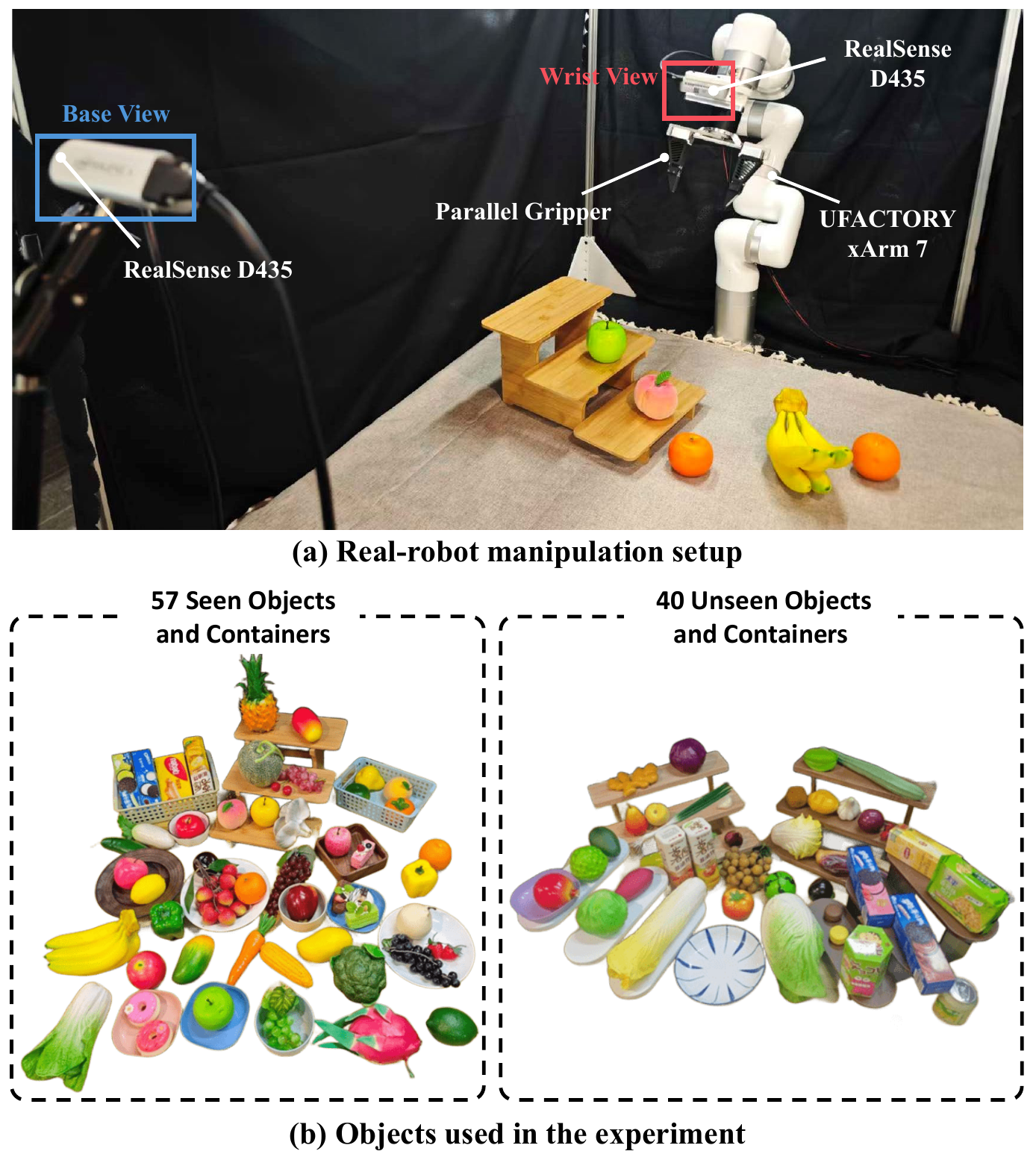}
\caption{\textbf{Real-robot system setup.} (a): Real-robot manipulation setup; (b): Overview of objects and containers used in real-robot language-guided manipulation. }
\label{fig:real_setup}
\end{figure}

\begin{figure*}[htbp]
\centering
% \vspace{-1em}
\includegraphics[width=0.95\linewidth]{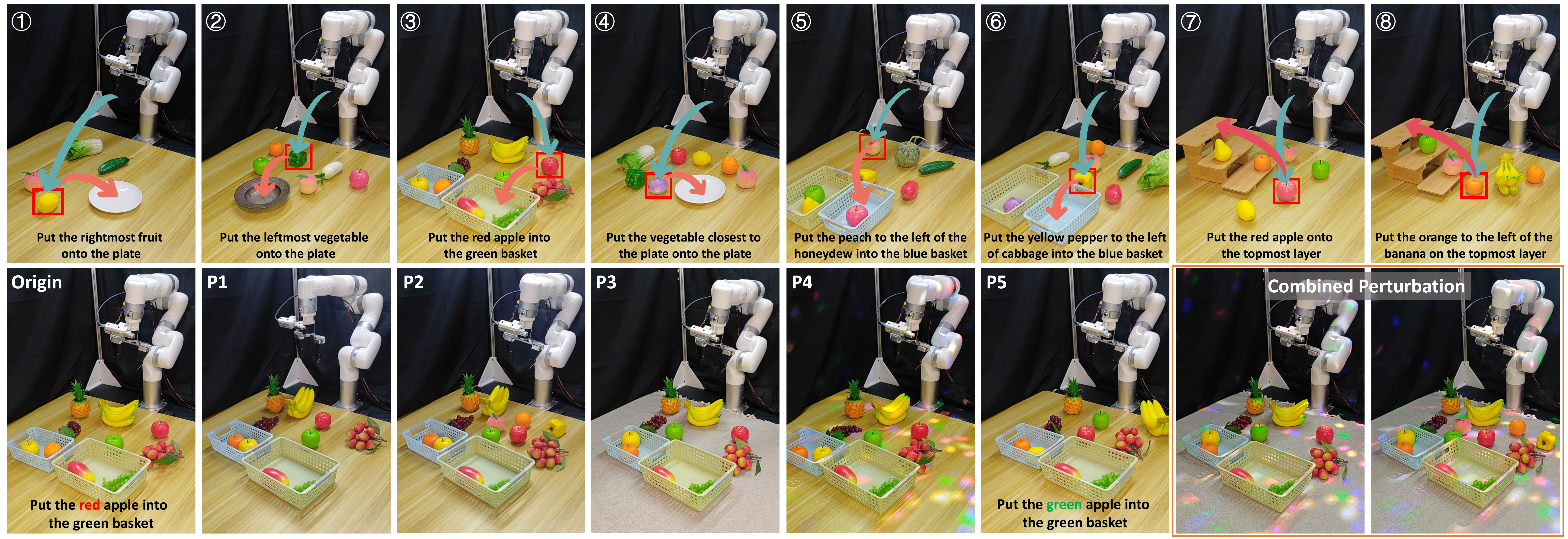}
\caption{\textbf{Real-robot tasks.} Top: Examples of scene setup and language instructions for 8 real robot tasks; Bottom: Original scenes, five types of perturbations (changes in robot initial pose, object interference, background changes, lighting perturbations, and variations in language instructions), along with examples of two combined perturbations. We showcase manipulation under combined perturbations in the multimedia demonstration, further illustrating the robustness of PokeVLA.
}
\vspace{3mm}
\label{fig:real_task}
\end{figure*}

\section{Real-World Experiments}

\subsection{Experiments Setup}
\subsubsection{Hardware Setup}
To conduct experiments and evaluate our method's performance in real-world scenarios, we set up a real-robot manipulation system as shown in Fig.~\ref{fig:real_setup} (a). The system consists of a UFACTORY xArm7 robotic arm equipped with a parallel gripper and two Realsense D435 cameras. Following the camera configurations in the simulator, one camera is mounted in front of the robot to provide a third-person view (base view), while the other is set at the end of the robotic arm (wrist view) to capture RGB observations.

\subsubsection{Data Collection}
% We assess the model across a range of spatial-reasoning manipulation tasks and their perturbation variants, including the original setting, variations in the robot’s initial pose, target-object disturbances, background perturbations, and lighting changes. For quantitative evaluation, we construct a real-world benchmark including everyday fruits, vegetables, and snacks, as shown in Fig.~\ref{fig:real_setup} (b). The benchmark consists of pick-and-place tasks within a 110 $\times$ 110 cm wooden tabletop workspace. It contains 57 seen objects and containers, and 40 unseen objects and containers. In each trial, the model must follow natural-language instructions to pick specified objects and place them into the correct container or location. To support post-training, we collect 50 hours of Gello teleoperated demonstrations exclusively using objects, containers, and shelf fixtures from the predefined “seen” set. We compare PokeVLA with OpenVLA-OFT and VLA-Adaptor under 5 evaluation settings.

We collected real-world robot demonstration data using an xArm 7 robotic arm equipped with the GELLO teleoperation system~\cite{wu2024gello}. The data consists of pick-and-place tasks performed within a 110 × 110 cm wooden tabletop workspace. To enhance the diversity of objects in the dataset and validate our method's instruction understanding, object tracking ability, and generalization capability in real-world scenarios, we included 97 different objects and containers. Among these, 57 objects and containers were previously encountered during data collection, while the remaining 40 unseen objects and containers serve as benchmarks. Additionally, during data collection, we deliberately varied object positions and arrangements to further increase diversity. Furthermore, the accompanying language instructions intentionally incorporate spatial referring expressions—such as left/right, front/back, and above/below—encompassing a wide range of spatial relationships between objects.

In total, we recorded 60 distinct tasks, each with 50 demonstrations, resulting in 3,000 trajectories. For each demonstration, synchronized RGB video streams were captured from the two camera viewpoints mentioned above (base view and wrist view).

\begin{table}[ht]
    \centering
    \caption{Real-world task list.}
    \label{tab:real tasks}
    \begin{tabular}{l p{0.68\linewidth}}
        \toprule
        \textbf{Task ID} & \textbf{Task Description} \\
        \midrule
        Task 1 & Put the rightmost fruit onto the plate. \\
        Task 2 & Put the leftmost vegetable onto the plate. \\
        Task 3 & Put the red apple into the green basket. \\
        Task 4 & Put the vegetable closest to the plate onto the plate. \\
        Task 5 & Put the peach to the left of the honeydew into the blue basket. \\
        Task 6 & Put the yellow pepper to the left of the cabbage into the blue basket. \\
        Task 7 & Put the red apple onto the topmost layer. \\
        Task 8 & Put the orange to the left of the banana on the topmost layer. \\
        \bottomrule
    \end{tabular}
\end{table}

\vspace{2mm}
\begin{table*}[htbp]
\caption{Comparison of real-world performance across different methods on eight tasks.}
\label{tab: real_1}
\centering
\renewcommand\arraystretch{1.2}
\begin{tabular}{lccccccccc} 
\toprule
Method & Task1 & Task2 & Task3 & Task4 & Task5 & Task6 & Task7 & Task8 & AVG \\
\midrule
OpenVLA-OFT & 3/10 & 2/10 & 4/10 & 3/10 & 0/10 & 2/10 & 1/10 & 1/10 & 20.00\% \\
VLA-Adapter & 7/10 & 8/10 & 8/10 & 9/10 & 4/10 & 8/10 & 6/10 & 5/10 & 68.75\% \\
PokeVLA & 9/10 & 9/10 & 8/10 & 9/10 & 7/10 & 7/10 & 8/10 & 8/10 & 81.25\% \\
\bottomrule
\end{tabular}
\label{tab:performance_comparison}
\end{table*}

\subsubsection{Data Annotation} 
We annotated the side-view images using a human-in-the-loop approach assisted by the SAM2 \cite{ravi2024sam} model. Specifically, we generated pixel-wise masks for the target object (to be manipulated) and the reference object mentioned in each instruction. These annotations provide high-quality supervision for vision–language–action modeling.

\subsubsection{Implement Details}
During the post-training phase, we train PokeVLA as well as the other two baselines, OpenVLA-OFT and VLA-Adapter, with the same settings. We train these models on the collected real-world demonstrations across 8 NVIDIA A100 GPUs for 50,000 iterations, with a per-GPU batch size of 4, resulting in an effective global batch size of 32.
For optimization, we adopt the AdamW optimizer and LoRA scheme. For training stability, we configure a learning rate of 1e-4 and utilize a cosine annealing learning rate scheduler with warm-up over the first 10\% of training steps.
Besides, we use the pretrained PokeVLM as a vision-language backbone. All models, including baselines, are trained and executed in real-world experiments using joint-space absolute control.

\subsection{Task Setting}
Similar to our simulation experiments, our real-world robot experiments are also conducted under two settings: the original setting and the perturbed setting, to assess the model's robustness. For the original setting, we evaluate the model on eight carefully designed tasks based on the collected dataset. The scene configurations and instruction details for each task are illustrated in Fig.~\ref{fig:real_task} and summarized in Table~\ref{tab:real tasks}. In evaluation, we introduce previously unseen objects as distractors (without violating the language instructions) and apply moderate variations to the target objects and scene layouts.

The eight tasks comprehensively evaluate the model’s capabilities: Task 1 tests spatial referencing, Tasks 2 and 4 assess both spatial and semantic understanding, Task 3 focuses on color referencing, Tasks 5 and 6 examine color and spatial referencing combined, while Tasks 7 and 8 impose stricter spatial demands—including left/right distinctions and vertical-level (layer) discrimination.

For the perturbed setting, we select Task 3 and Task 8. As shown in Fig.~\ref{fig:real_task}, we introduce five types of perturbations on top of the original scenarios: changes in the robot’s initial pose, object interference, background variations, lighting fluctuations, and modifications to language instructions.

Additionally, in the accompanying multimedia demonstration, we showcase manipulation under various combined perturbations, further illustrating the robustness of PokeVLA.

\subsection{Quantitative Results}

Table~\ref{tab: real_1} presents the success rates of the three methods across eight tasks under the original setting. Compared to the two baselines, our method demonstrates a clear performance improvement. Compared to VLA-Adapter, PokeVLA demonstrates a significant advantage in Task 1, as well as in the more challenging Tasks 7 and 8, thereby proving our method's superior spatial understanding capabilities. This highlights the effectiveness of enhancing spatial grounding during VLM pre-training and geometry alignment during post-training. Additionally, our method performs better in Task 5, indicating that the VLM pre-training has improved our approach's ability to understand color references.

Table~\ref{tab: real_2} displays the performance under five different perturbations. Compared to the model of comparable scale, VLA-Adapter, PokeVLA improves the success rate by 20\%. Across every scene perturbation in the two tasks, PokeVLA achieves the best performance. Notably, its advantage is most pronounced under robot initial pose and lighting disturbances, benefiting from PokeVLA's robust learning of manipulation-related representations.

Experiments in the real world further validate the effectiveness and robustness of PokeVLA.

\subsection{Qualitative Results}
\begin{figure*}[!ht]
\centering
\includegraphics[width=0.99\textwidth]{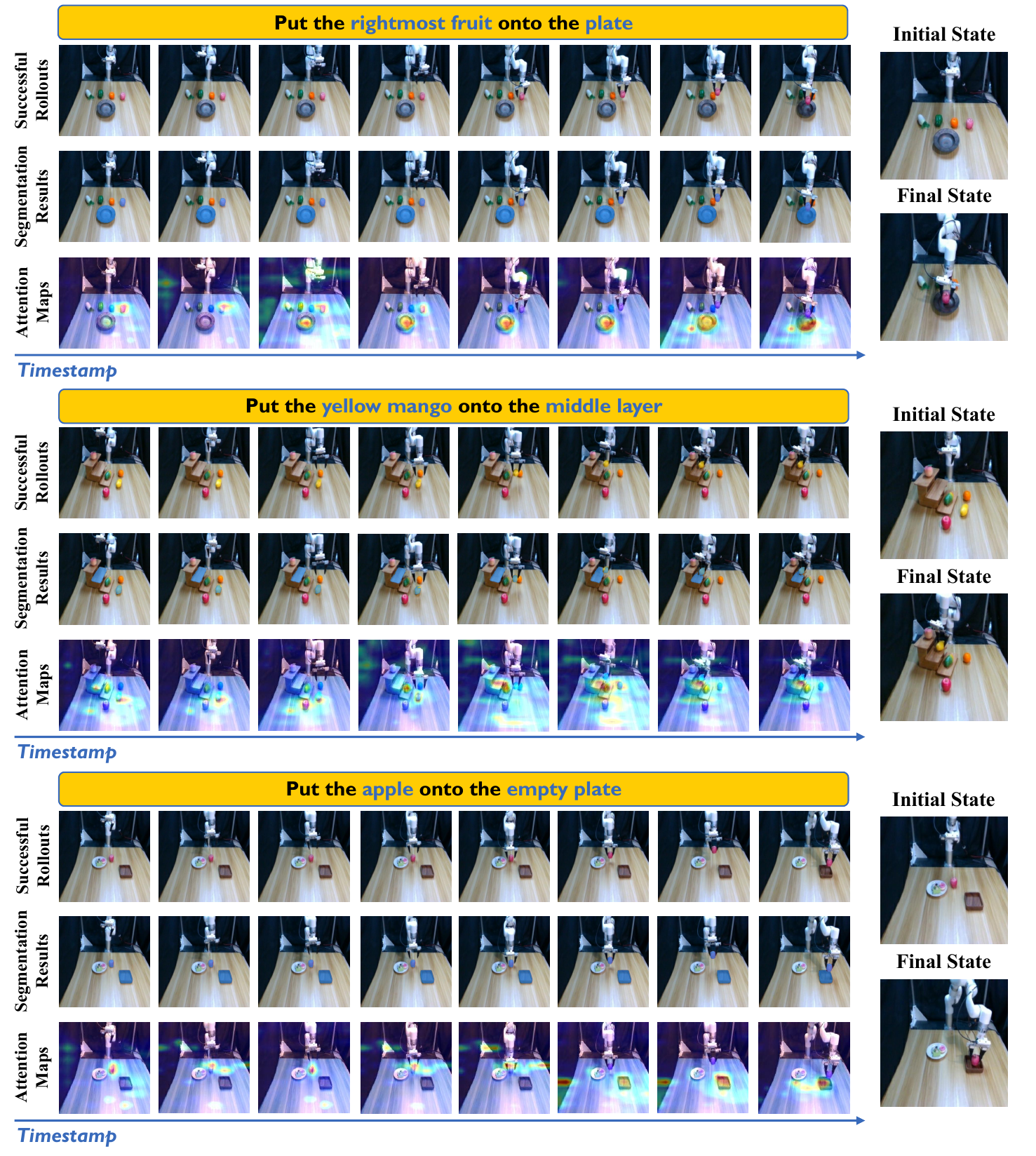}
% \vspace{-2em}
\caption{\textbf{Visualization of real-world experiments.} Our model can generalize to complex instructions involving spatial and color referencing on real-world tasks, showing superior semantic understanding and task performance while consistently maintaining accurate manipulation target segmentation and goal-oriented attention over long time horizons. Best viewed zoom in.}
\label{fig:real_visualization}
\end{figure*}

As shown in Figure~\ref{fig:real_visualization}, we present visualizations of three manipulation tasks designed to test the model's comprehension of spatial relations, color referring, and semantic understanding. (For instance, the first task requires identifying the "rightmost" object and distinguishing between fruits and vegetables; the second task demands distinguishing between two mangoes based on color ["yellow"]; the third task involves understanding the state of a plate ["empty"]). We visualize the manipulation process alongside the results of goal-aware segmentation and the model's attention maps.
The results demonstrate that our model consistently performs accurate scene understanding and executes corresponding actions, whether the instruction involves spatial referring (e.g., "rightmost" in the first row, "middle" in the second row), color referring (e.g., "yellow" in the second row), or state of objects (e.g., "empty" in the third row), showcasing its robust semantic understanding. Furthermore, our auxiliary task, goal-aware segmentation learning, maintains superior performance even in real-world environments. They provide consistent and accurate segmentation results throughout long-horizon tasks and effectively guide the model's attention to focus on task-relevant regions, contributing to successful task completion.

Overall, these results highlight the superiority of our approach in instruction following, spatial understanding, and target focusing, demonstrating its strong generalization capability in real-world scenarios.

% \begin{table*}
% \caption{Comparison of real-world performance under different perturbation. P1–P5 denote: (P1) end-effector initial pose perturbation, (P2) object perturbation, 
% (P3) background perturbation (P4) lighting perturbation, and (P5) unseen language instructions.}
% \centering
% \renewcommand\arraystretch{1.2}
% \begin{tabular*}{0.9\textwidth}
% {@{\extracolsep{\fill}}l c c c c c c c c c c c c c c} 

% \toprule
% \multirow{2}{*}{Method} & \multicolumn{6}{c}{Task3} & \multirow{2}{*}{AVG (\%)} & \multicolumn{6}{c}{Task8} & \multirow{2}{*}{AVG (\%)} \\
%  & Origin & P1 & P2 & P3 & P4 & P5 & & Origin & P1 & P2 & P3 & P4 & P5 & \\
% \midrule
% OpenVLA-OFT & 4/10 & 3/10 & 1/10 & 0/10 & 0/10 & 0/10 & 8\% & 1/10 & 0/10 & 0/10 & 0/10 & 0/10 & 0/10 & 2\% \\
% \midrule
% VLA-Adapter & 8/10 & 6/10 & 7/10 & 4/10 & 7/10 & 4/10 & 56\% & 5/10 & 2/10 & 4/10 & 1/10 & 4/10 & 4/10 & 30\% \\
% \midrule
% PokeVLA & 8/10 & 8/10 & 7/10 & 6/10 & 8/10 & 6/10 & 70\% & 8/10 & 7/10 & 5/10 & 3/10 & 7/10 & 6/10 & 56\% \\
% \bottomrule
% \end{tabular*}
% \label{tab:robustness_comparison}
% \end{table*}

\begin{table*}
\caption{Comparison of real-world performance under different perturbations. P1–P5 denote: (P1) end-effector initial pose perturbation, (P2) object perturbation, (P3) background perturbation, (P4) lighting perturbation, and (P5) unseen language instructions. We provide the success rate of the original setting for reference, which is excluded from the average calculation.}
\label{tab: real_2}
\centering
\renewcommand\arraystretch{1.2}
\begin{tabular*}{0.98\textwidth}{@{\extracolsep{\fill}}l *{7}{c} *{7}{c} c}
\toprule
\multirow{2}{*}{Method} &
\multicolumn{7}{c}{Task3} &
\multicolumn{7}{c}{Task8} &
\multirow{2}{*}{Overall AVG} \\
\cmidrule(lr){2-8} \cmidrule(lr){9-15}
& \cellcolor{lightgray}Origin & P1 & P2 & P3 & P4 & P5 & AVG 
& \cellcolor{lightgray}Origin & P1 & P2 & P3 & P4 & P5 & AVG & \\
\midrule
OpenVLA-OFT 
& \cellcolor{lightgray}4/10 & 3/10 & 1/10 & 0/10 & 0/10 & 0/10 & 8.0\%
& \cellcolor{lightgray}1/10 & 1/10 & 0/10 & 0/10 & 0/10 & 0/10 & 2.0\% & 5.0\% \\
\midrule
VLA-Adapter 
& \cellcolor{lightgray}8/10 & 6/10 & 7/10 & 4/10 & 7/10 & 4/10 & 56.0\% 
& \cellcolor{lightgray}5/10 & 2/10 & 4/10 & 1/10 & 4/10 & 4/10 & 30.0\% & 43.0\% \\
\midrule
PokeVLA 
& \cellcolor{lightgray}8/10 & 8/10 & 7/10 & 6/10 & 8/10 & 6/10 & 70.0\%
& \cellcolor{lightgray}8/10 & 7/10 & 5/10 & 3/10 & 7/10 & 6/10 & 56.0\% & 63.0\% \\
\bottomrule
\end{tabular*}
\label{tab:robustness_comparison}
\end{table*}

\section{Conclusion}
In this work, we introduced PokeVLA, a tiny-scale yet powerful foundation model for embodied manipulation. To address key bottlenecks in prior VLA research, such as domain gaps from general-purpose VLMs, a lack of spatial consistency, and insufficient high-level task guidance, we developed a novel two-stage framework. This framework first pre-trains a compact, embodied-aware VLM (PokeVLM) on a large-scale curated dataset, then efficiently injects task-relevant spatial and semantic representations into the action learning process via multi-view geometric alignment and action queries.

Extensive evaluations on challenging benchmarks (LIBERO and LIBERO-Plus) and real-world tasks demonstrate that PokeVLA achieves state-of-the-art performance and superior generalization, even with only 1.22B parameters. Our method significantly outperforms larger or comparable baselines under challenging perturbations, showcasing robust scene understanding and instruction-following capability.
% \newpage
\bibliographystyle{IEEEtran}
\bibliography{reference}

\end{document}